\documentclass[10pt,twocolumn]{article}
\usepackage{graphicx}
\usepackage{amsmath,amssymb,times,cite}
\usepackage{xcolor}
\usepackage{booktabs}

\newcommand{\bg}[1]{\boldsymbol{#1}} 
\newcommand{\bm}[1]{\mathbf{#1}} 

\newcommand\T{{\mathpalette\raiseT\intercal}}
\newcommand\raiseT[2]{%
\setbox0\hbox{$#1{#2}$}\raise\dp0\box0}

\title{\LARGE\textbf{Ridge Regression Neural Network for Pediatric Bone Age Assessment}}
\author{Ibrahim Salim and A. Ben Hamza\\
Concordia Institute for Information Systems Engineering\\
Concordia University, Montreal, QC, Canada
}
\date{}

\begin{document}
\maketitle

\begin{abstract}
Bone age is an important measure for assessing the skeletal and biological maturity of children. Delayed or increased bone age is a serious concern for pediatricians, and needs to be accurately assessed in a bid to determine whether bone maturity is occurring at a rate consistent with chronological age. In this paper, we introduce a unified deep learning framework for bone age assessment using instance segmentation and ridge regression. The proposed approach consists of two integrated stages. In the first stage, we employ an image annotation and segmentation model to annotate and segment the hand from the radiographic image, followed by background removal. In the second stage, we design a regression neural network architecture composed of a pre-trained convolutional neural network for learning salient features from the segmented pediatric hand radiographs and a ridge regression output layer for predicting the bone age. Experimental evaluation on a dataset of hand radiographs demonstrates the competitive performance of our approach in comparison with existing deep learning based methods for bone age assessment.
\end{abstract}


\section{Introduction}
Bone age assessment is a fundamental problem in the diagnosis and treatment of children and adolescents with suspected growth disorder~\cite{Gilsanz:12}. It is often used in clinical practice by pediatricians for the children's skeletal maturation assessment in an effort to determine the difference between a child's bone age and a chronological age~\cite{Martin:11}. This difference influences the decisions taken by clinicians to predict the child's accurate age and often leads to errors in managing the diagnosis of children with skeletal dysplasias, and metabolic and endocrine disorders~\cite{Satoh:15}.

The assessment of bone age has been traditionally performed using the Greulich-Pyle (GP) and Tanner-Whitehouse (TW) methods~\cite{Greulich:59,Tanner:75}, which are based on left hand and wrist radiographs. In the GP method, the bones in the hand and wrist radiographs are compared to the bones of a standard atlas, while the TW method relies on a scoring system that examines the level of skeletal maturity for twenty selected regions of interest in specific bones of the hand-wrist and a numerical score is then assigned to selected hand-wrist bones depending on the appearance of certain well-defined maturity indicators. However, these manual methods suffer from substantial inter- and intra-observer variability, rely on trained radiologists, and are time consuming~\cite{somkantha:11,Alshamrani:19}. To circumvent these issues, several automated bone age assessment approaches based on image processing and computer vision techniques have been proposed, including BoneXpert~\cite{Thodberg:09}, which makes use of conventional radiographs of the hand according to the GP and TW methods. While BoneXpert is an automatic method, it does, however, require some manual analysis, particularly for X-ray images with low quality.

The recent trend in bone age assessment is geared toward automated methods using deep neural networks to learn features from hand radiographs at various levels of abstraction. This trend has been driven, in part, by a combination of affordable computing hardware, open source software, and the availability of large-scale datasets~\cite{Bengio:09,Schmidhuber:15,Halabi:19}. Several deep learning based models have been recently proposed to tackle the bone age assessment problem~\cite{spampinato:17,lee:17,larson:17,tong:18,van:18,iglovikov:18,Wu:19,Chen:19,Liu:19}, achieving good performance with substantially improved results. Spampinato \textit{et al.}~\cite{spampinato:17} presented a BoNet architecture comprised of five convolutional layers, one deformation layer, one fully connected later, and a linear scalar layer. They showed that BoNet outperforms fine-tuned convolutional neural networks for assessing bone age over races, age ranges and gender. Lee \textit{et al.}~\cite{lee:17} designed a deep learning model to detect and segment the hand and wrist prior to performing bone age assessment with a fine-tuned convolutional neural network. Larson \textit{et al.}~\cite{larson:17} applied a fifty-layer residual network to estimate bone age, achieving comparable performance to that of trained human reviewers. However, their model is not effective at predicting the bone age of patients younger than two years.

In this paper, we propose an integrated deep learning based framework, called RidgeNet, for bone age assessment using instance segmentation and ridge regression. We develop a regression network architecture for bone age assessment using a pre-trained deep learning model in conjunction with a regularized regression output layer. The main contributions of this paper can be summarized as follows:
\begin{itemize}
\item We present an image annotation and segmentation model to annotate and segment the hand from the radiographic image with a minimum number of annotated images, followed by background removal.
\item We leverage the power of transfer learning with fine-tuning to learn salient features from the segmented radiographs.
\item We design a regression neural network architecture with a ridge regression output layer for predicting the bone age.
\item We show through extensive experiments the competitive performance of the proposed approach in comparison with baseline methods.
\end{itemize}

The rest of this paper is organized as follows. In Section 2, we review important relevant work. In Section 3, we introduce a two-stage approach for bone age assessment using instance segmentation and ridge regression. In the first stage, we employ an image annotation and instance segmentation model to extract and separate different regions of interests in an image. In the second stage, we leverage the power of transfer learning by designing a deep neural network with a ridge regression output layer. Section 4 presents experimental results to demonstrate the competitive performance of our approach compared to baseline methods. Finally, Section 5 concludes the paper and points out future work directions.

\section{Related Work}
The basic goal of bone age assessment is to evaluate the biological maturity of children. To achieve this goal, various bone age prediction methods based on deep learning have been proposed. Tong \textit{et al.}~\cite{tong:18} presented an automated skeletal bone age assessment model using convolutional neural networks and support vector regression with multiple kernel learning by combining heterogeneous features from X-ray images, race and gender. Van Steenkiste~\textit{et al.}~\cite{van:18} integrated a pre-trained convolutional neural network with Gaussian process regression in order to refine the estimated bone age by exploiting variations in
the predictions. The Gaussian process regression is used to estimate a vector of prediction scores for rotated and mirror images of a single radiograph. Iglovikov \textit{et al.}~\cite{iglovikov:18} proposed deep learning based regression and classification models using convolutional neural networks by applying image segmentation via U-Net, a fully convolutional network that extracts regions of interest (ROIs) from images and predicts each pixel's class~\cite{ronneberger:15}. With the exception of the last two layers, the regression model proposed in~\cite{ronneberger:15} is similar to the classification one, where each bone age is assigned a class. In the last layer of the classification model, probabilities obtained from the softmax layer are multiplied by a vector of bone ages uniformly distributed over integer values. Wu \textit{et al.}~\cite{Wu:19} designed a network architecture consisting of an instance segmentation model and a residual attention network. The instance segmentation model segments the hands from X-ray images to avoid the distractions of other objects, while the residual attention network forces the network to focus on the main components of the X-ray images. Liu \textit{et al.}~\cite{Liu:19} replaced the encoder of U-Net with a pre-trained convolutional neural network to perform image segmentation, followed by applying a ranking learning technique instead of regression to assess bone age. Similarly, Pan \textit{et al.}~\cite{pan:20} introduced a U-Net based model, which consists of image segmentation, feature extraction, and ensemble modules. More recently, Liu \textit{et al.}~\cite{Liu:20} proposed a bone age assessment model, which is trained on multiclassifiers based on ensemble learning, to predict the optimal segmentation threshold for hand mask segmentation. In~\cite{Wibisono:20}, a region-based feature connected layer from the essential segmented region of a hand X-ray is introduced in order to predict bone age using deep learning models.

While these deep learning based approaches have yielded competitive results in bone age assessment, they suffer, however, from high model complexity, often require a pre-processing image alignment step, and do not distinguish between important from less-important predictors in the regression model. In addition, the U-Net architecture and its variants suffer from the large semantic gap between the low- and high-level features of the encoder and decoder subnetworks, leading to fusing semantically dissimilar features and hence resulting in blurred feature maps throughout the learning process and also adversely affecting the output segmentation map by under- and/or over-segmenting regions of interest (ROIs).

\section{Method}
Bone age prediction is typically achieved by extracting ROI features from pediatric hand images, followed by using a learning model to estimate the bone age of these radiographs. Our algorithm is divided into two stages. In the first stage, we carried out an image preprocessing step for radiographs using image annotation and segmentation. In the second stage, we trained the proposed deep learning model using the segmented radiographs to evaluate the proposed model's performance.

\subsection{Image Annotation and Segmentation Model}
This preprocessing step aims at extracting regions of interest from radiographs using image annotation and instance segmentation.

\begin{figure}[!htb]
\centering
\includegraphics[scale=.23]{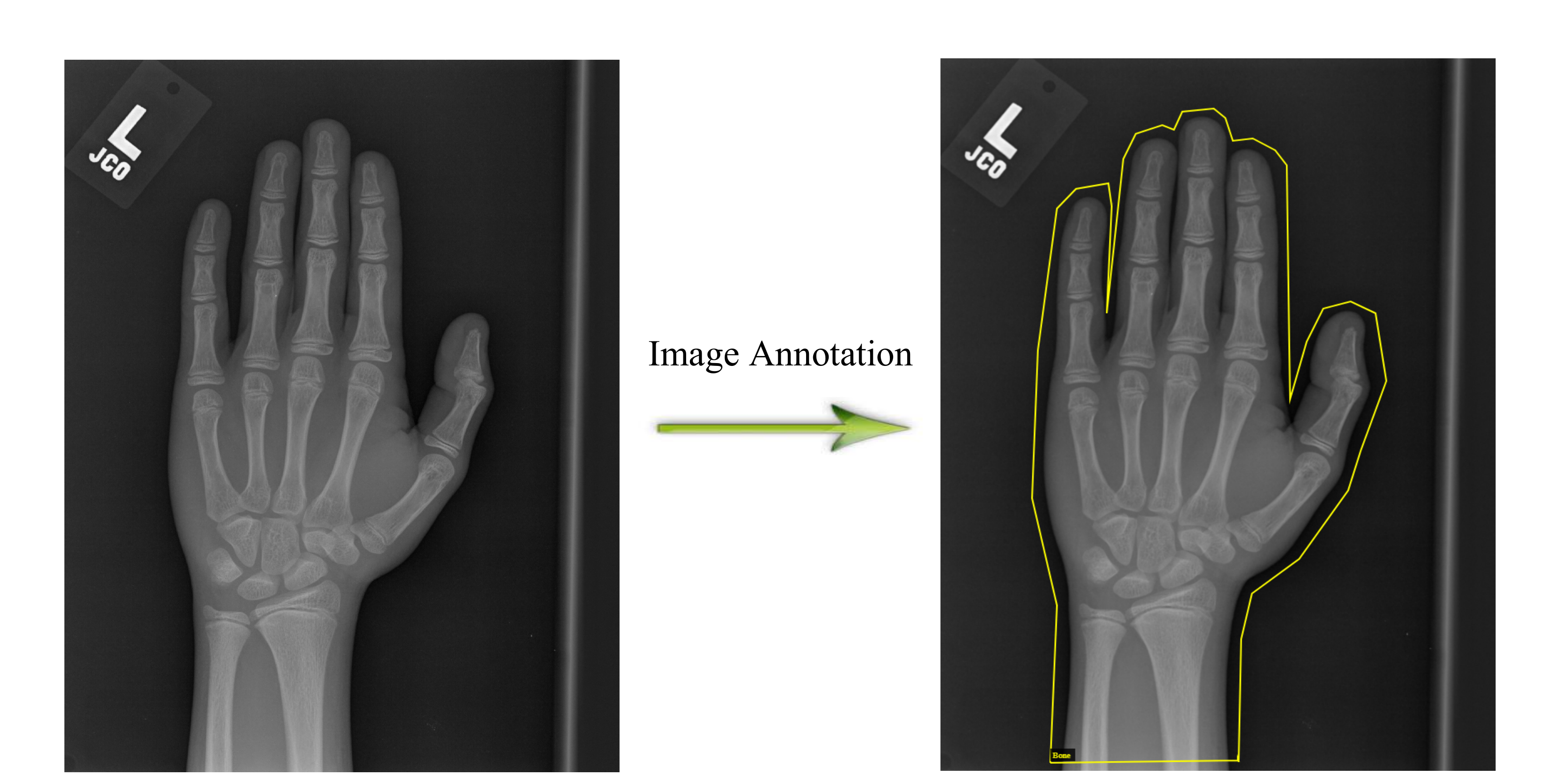}
\caption{Image annotation using VGG image annotator.}
\label{Fig:Annotated_Image}
\end{figure}

\medskip
\noindent{\textbf{Image Annotation:}}\quad We use the VGG image annotator\footnote{http://www.robots.ox.ac.uk/~vgg/software/via/via-2.0.4.html} to define regions in an image and create textual descriptions of those regions for image segmentation purposes. The annotations can be manually done using rectangles, circles, ellipses, polygons, lines, or points, and then converted to common objects in context (COCO) dataset format, which is the commonly used format for object detection and instance segmentation algorithms. As shown in Figure~\ref{Fig:Annotated_Image}, the annotated hand is obtained using a polygonal region shape.

\begin{figure}[!htb]
\centering
\includegraphics[width=3.6in,height=2.6in]{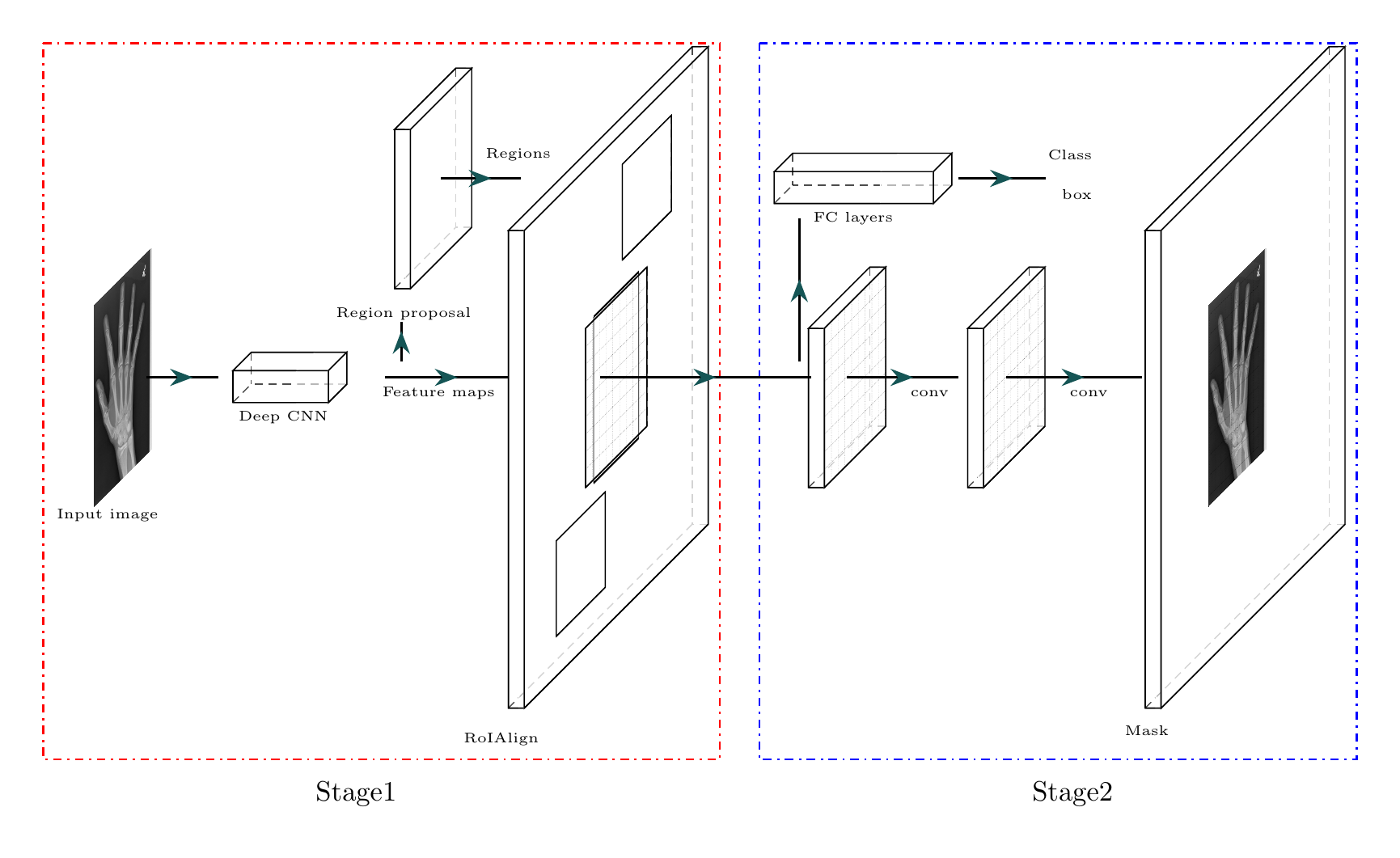}
\caption{Instance segmentation using Mask R-CNN.}
\label{Fig:Mask_R-CNN_Network}
\end{figure}

\medskip
\noindent{\textbf{Instance Segmentation:}}\quad Instance segmentation refers to the task of detecting and delineating each distinct object of interest appearing in an image, and includes identification of boundaries of the objects at the detailed pixel level. Several instance segmentation techniques have been recently proposed, including mask region-based convolutional neural network (Mask R-CNN), which locates each pixel of every object in the image instead of the bounding boxes~\cite{he:17}. The input to the Mask R-CNN algorithm is an image and the output is a bounding box and a mask that segment each object in the image, as shown in Figure~\ref{Fig:Mask_R-CNN_Network}. Mask R-CNN consists of two main stages. The first stage scans the image and generates proposals about the regions where there might be an object. The second stage classifies the proposals and generates bounding boxes and masks. These two stages are connected to a backbone network that uses residual neural network in conjunction with feature pyramid network for feature extraction. The feature pyramid network constructs a high-level feature pyramid and recognizes objects at different scales. In addition to using a region-of-interest (RoI) pooling, which performs max pooling on inputs of non-uniform sizes and produces a small feature map of fixed size, Mask R-CNN employs an ROIAlign layer that aligns the extracted features with the input in a bid to avoid any quantization of the RoI boundaries or bins.

\begin{figure}[!htb]
\centering
\includegraphics[scale=.2]{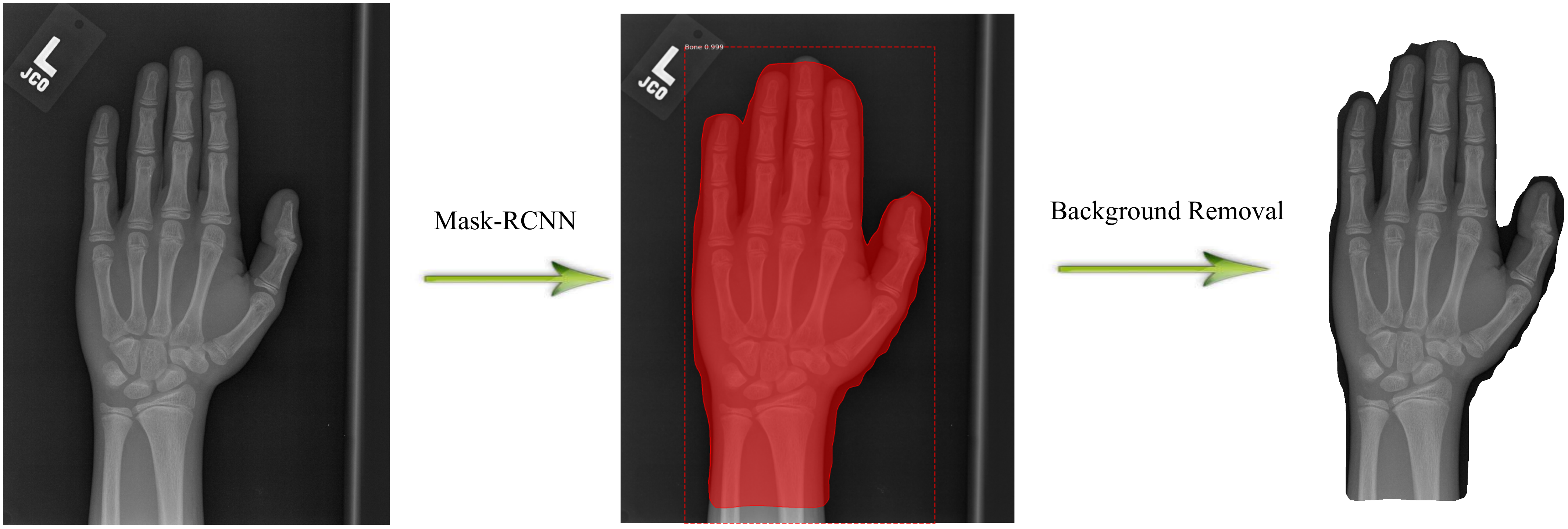}
\caption{Image segmentation using Mask R-CNN, followed by background removal.}
\label{Fig:Image_Segmentation}
\end{figure}

For radiograph segmentation, we leverage Mask R-CNN that was trained on the COCO dataset. The given annotated images are split into two disjoint subsets: the training set for learning, and the test set for testing. We used the pre-trained weights of some layers of Mask R-CNN as initial weights. More specifically, we trained the region proposal network, classifier and mask heads layers of Mask R-CNN and freezed the other hidden layers to speed-up the training of the network. The trainable weights generated after training are used to perform instance segmentation. The flowchart for image segmentation and background removal is depicted in Figure~\ref{Fig:Image_Segmentation}. As can be seen, the detected part of the image (whole hand) is classified as a foreground with a very high accuracy, while the rest of the radiograph image is classified as a background. Note that the red colored hand and the dotted rectangle represent the bounding box and mask, respectively.

\subsection{Bone Age Assessment Model}
\begin{figure*}[]
\centering
\includegraphics[scale=0.32]{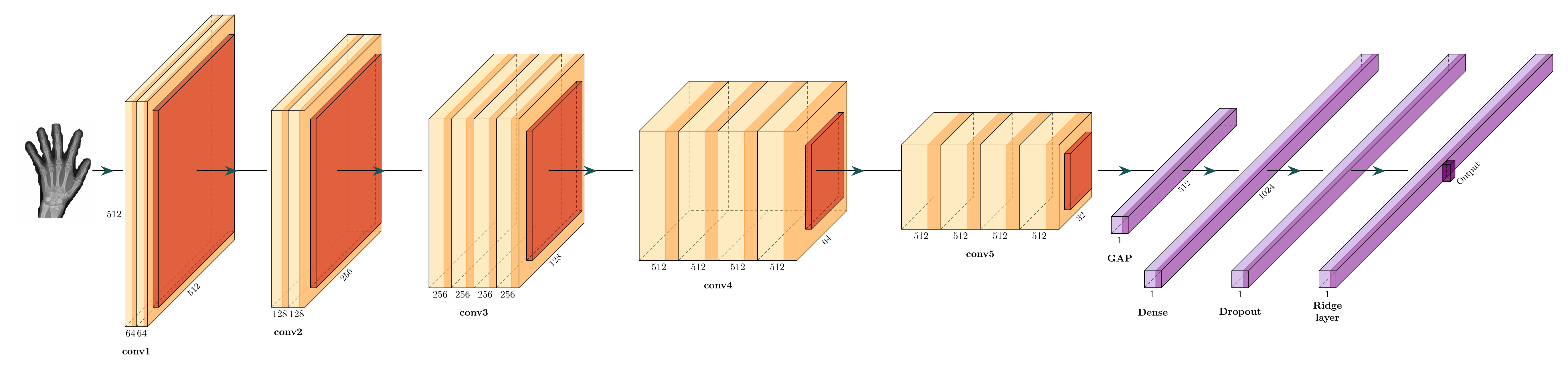}
\caption{Architecture of the proposed regression network.}
\label{Fig:Proposed_Pre-trained_CNN}
\end{figure*}

The proposed bone age assessment model uses the pre-trained VGG-19 convolutional neural network with a ridge regression output layer, as illustrated in Figure~\ref{Fig:Proposed_Pre-trained_CNN}. The VGG-19 network consists of 19 layers with learnable weights: 16 convolutional layers, and 3 fully connected layers~\cite{simonyan:14}. Each convolutional layer uses a $3\times 3$ kernel with stride 2 and padding 1. A $2\times 2$ max-pooling is performed with stride 2 and zero padding. As shown in Figure~\ref{Fig:Proposed_Pre-trained_CNN}, the proposed architecture consists of five blocks of convolutional layers, followed by a global average pooling (GAP), a fully connected layer of 1024 neurons, a dropout layer, a ridge regression layer, and single regression output. Each of the first and second blocks is comprised of two convolutional layers with 64 and 128 filters, respectively. Similarly, each of the third, fourth and fifth blocks consists of four convolutional layers with 256, 512, and 512 filters, respectively. The GAP layer, which is widely used in object localization, computes the average output of each feature map in the previous layer and helps minimize overfitting by reducing the total number of parameters in the model. GAP turns a feature map into a single number by taking the average of the numbers in that feature map, and helps identify where deep neural networks pay attention. Similar to max pooling layers, GAP layers have no trainable parameters and are used to reduce the spatial dimensions of a three-dimensional tensor.

Ridge regression, on the other hand, is a regularized regression method that shrinks the estimated coefficients towards zero. More specifically, given a response vector $\bm{y}\in\mathbb{R}^n$ and a predictor matrix $\bm{X}\in\mathbb{R}^{n\times p}$, the ridge regression coefficients are defined as
\begin{equation}
\hat{\bg{\beta}} = \arg\min_{\bg{\beta}\in\mathbb{R}^p}\left\{\|\bm{y}-\bm{X}\bg{\beta}\|_2^2+\lambda\|\bg{\beta}\|_2^2\right\},\quad \lambda\geq 0
\label{Eq:Ridge_Regression}
\end{equation}
where $\lambda$ is a tuning parameter that controls the strength of the penalty term and decides the shrinkage amount of the coefficients. A larger value of $\lambda$ indicates more shrinkage. In practice, the hyperparameter $\lambda$ is often estimated using cross-validation.

The solution to the ridge regression equation \eqref{Eq:Ridge_Regression} is given by
\begin{equation}
\hat{\bg{\beta}} = (\bm{X}^{\T}\bm{X}+\lambda \bm{I})^{-1} \bm{X}^{\T}\bm{y},
\label{Eq:Ridge_Regression_Matrix}
\end{equation}
which reduces to the linear regression estimate when $\lambda=0$. Unlike linear regression which does not differentiate ``important'' from ``less-important'' predictors in a model, ridge regression reduces model complexity and prevents over-fitting by producing new estimators that are shrunk closer to the ``true'' parameters.

The ridge regression layer computes the mean square error (MSE) given by
\begin{equation}
\text{MSE} = \frac{1}{n}\sum_{i=1}^{n}(y_i - \hat{y}_i)^2,
\label{Eq:MSE}
\end{equation}
where $n$ is the number of responses, $y_i$ is the target output, and $\hat{y}_i$ is the network's prediction for response $i$.

The goal of the ridge regression layer is to find the optimal values of the regression coefficients, while avoiding the multicollinearity issue that frequently occurs in multiple linear regression. Multicollinearity in a regression model is a statistical phenomenon that arises when some predictor variables in the model are correlated with other predictor variables, leading to increased variance of the regression coefficients and hence making them unstable. In order to mitigate the multicollinearity issue, ridge regression is usually used not only to reduce model complexity, but also to prevent over-fitting by adding a regularization term in an effort to ensure a smaller variance in resulting parameter estimates. Motivated by these nice properties of ridge regression, we design a ridge regression layer and incorporate it into our proposed network architecture with the goal of circumventing both the multicollinearity and over-fitting issues.

\section{Experiments}
In this section, we conduct extensive experiments to evaluate the performance of the proposed RidgeNet model in bone age assessment. The effectiveness of RidgeNet is validated by performing a comprehensive comparison with several baseline methods.

\begin{figure}[!htb]
\setlength{\tabcolsep}{.15em}
\centering
\begin{tabular}{ccc}
\includegraphics[width=1.1in,height=1.2in]{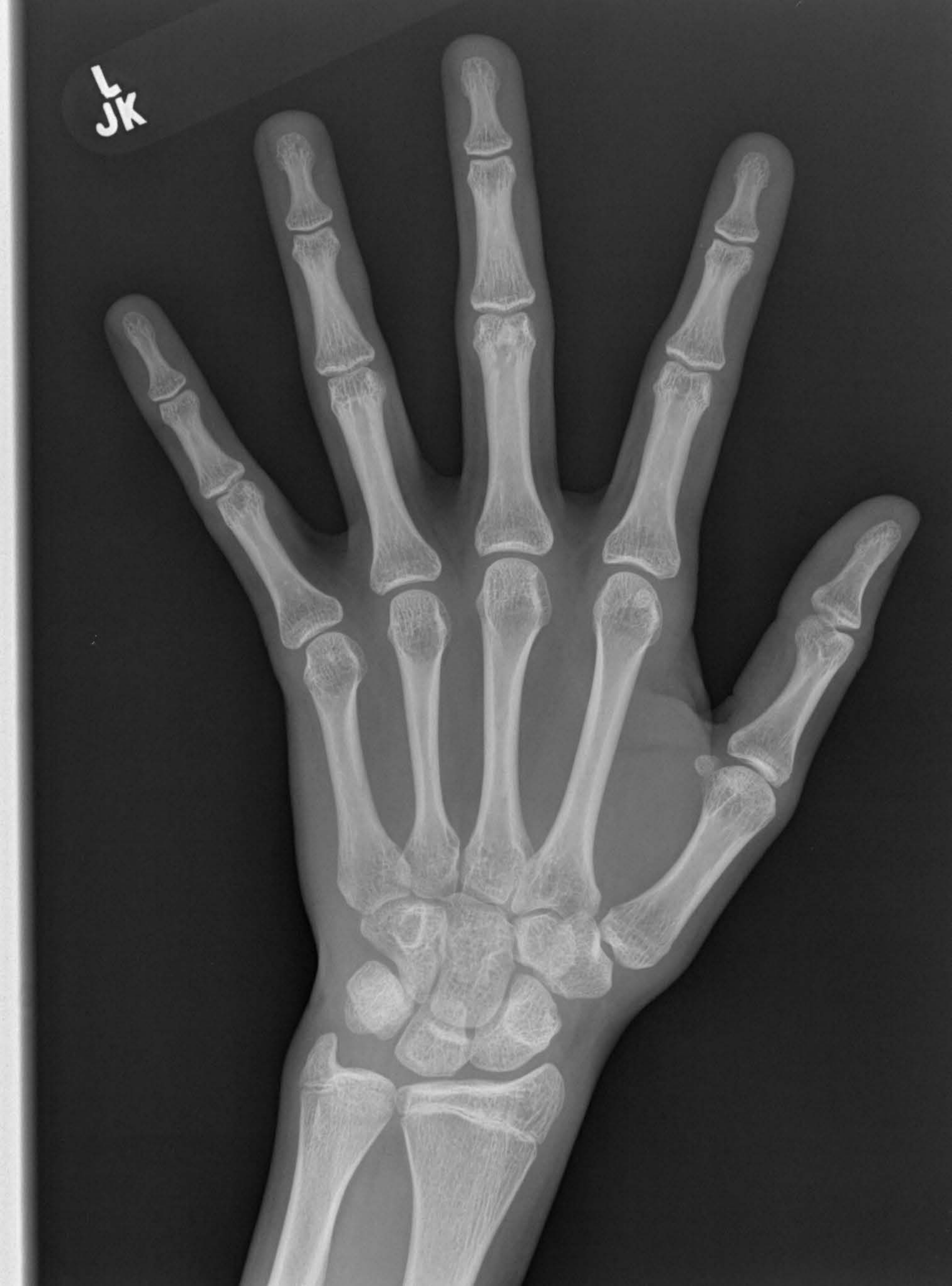}&
\includegraphics[width=1.1in,height=1.2in]{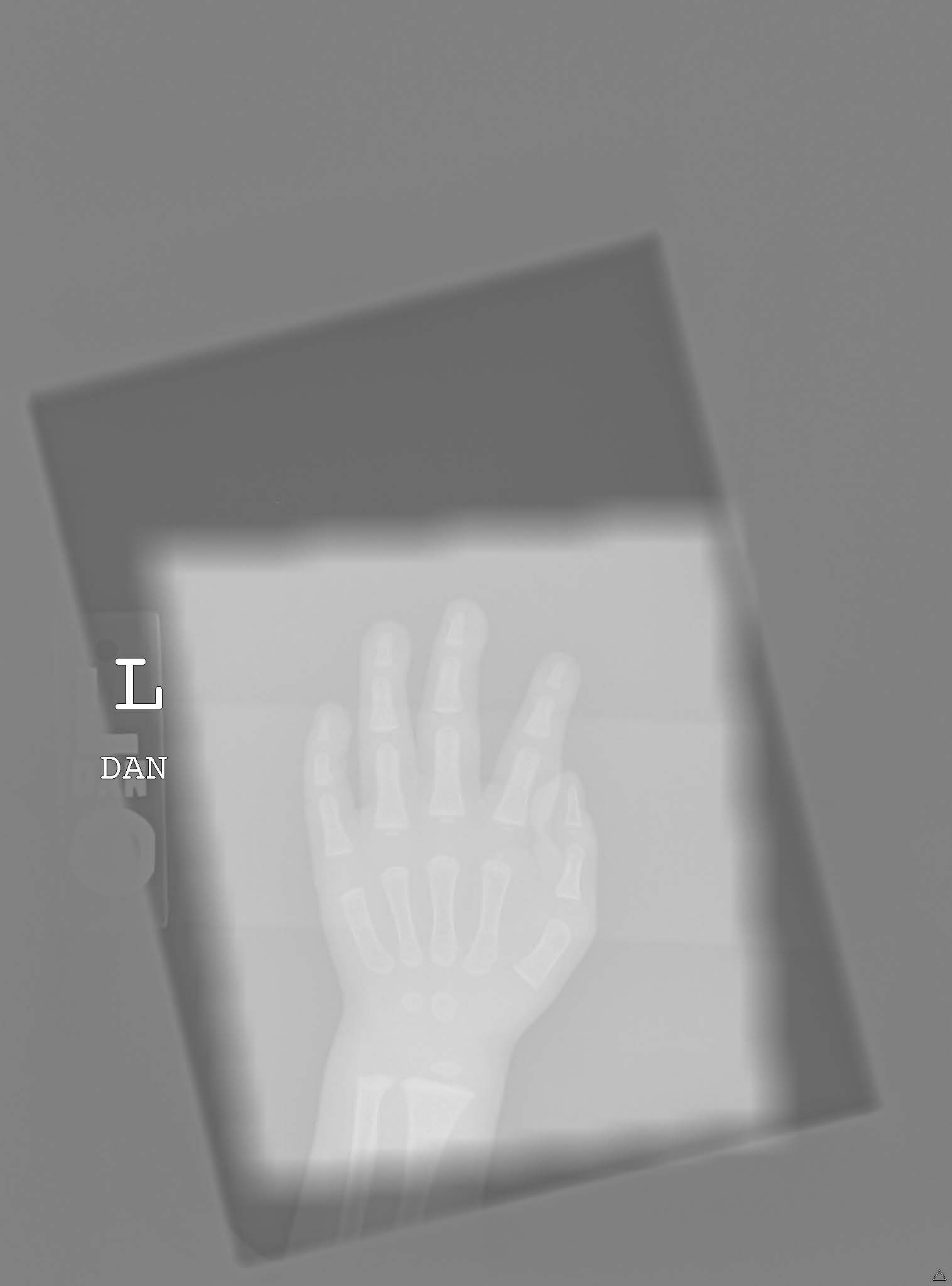}&
\includegraphics[width=1.1in,height=1.2in]{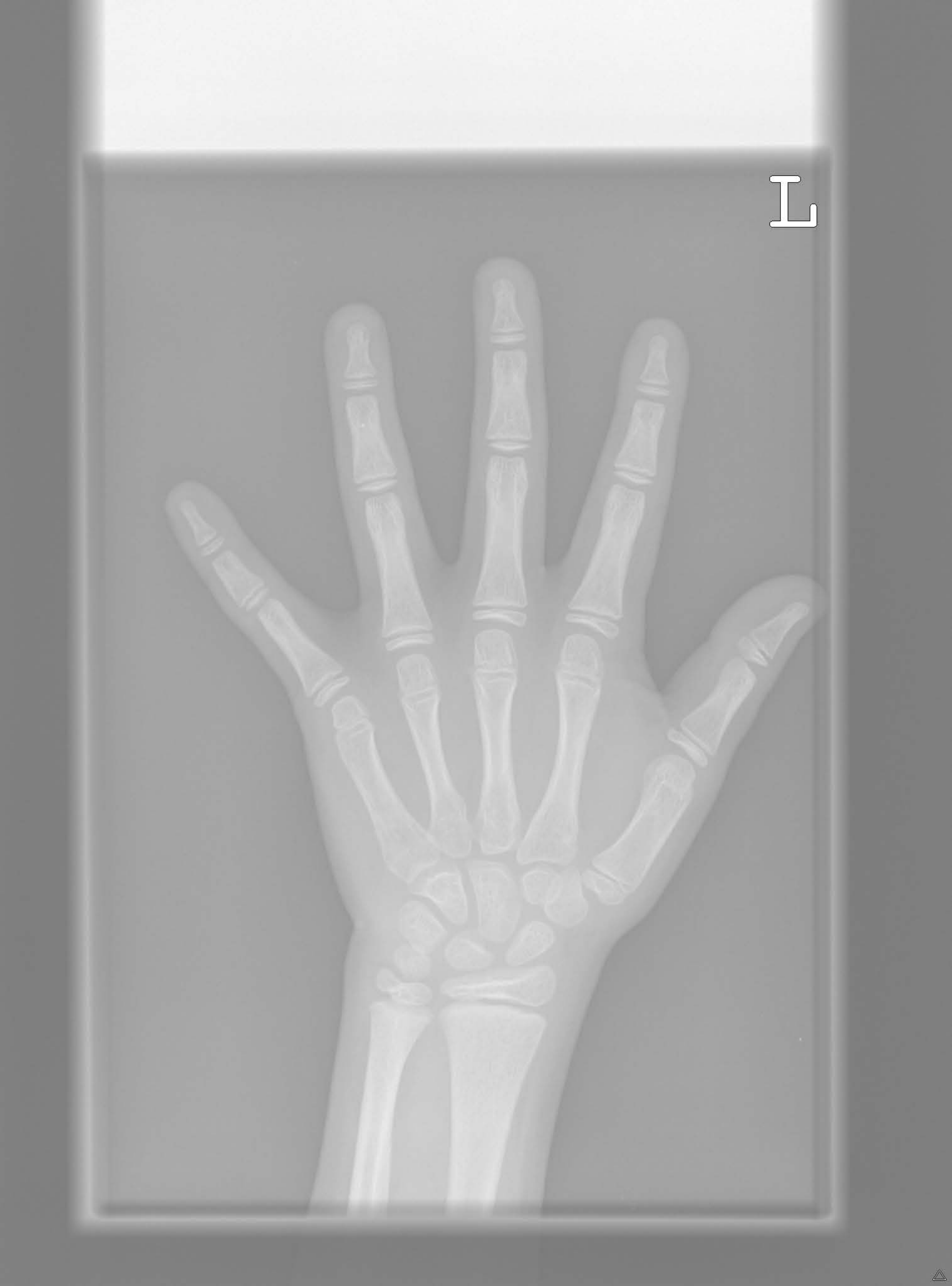}\\
\includegraphics[width=1.1in,height=1.2in]{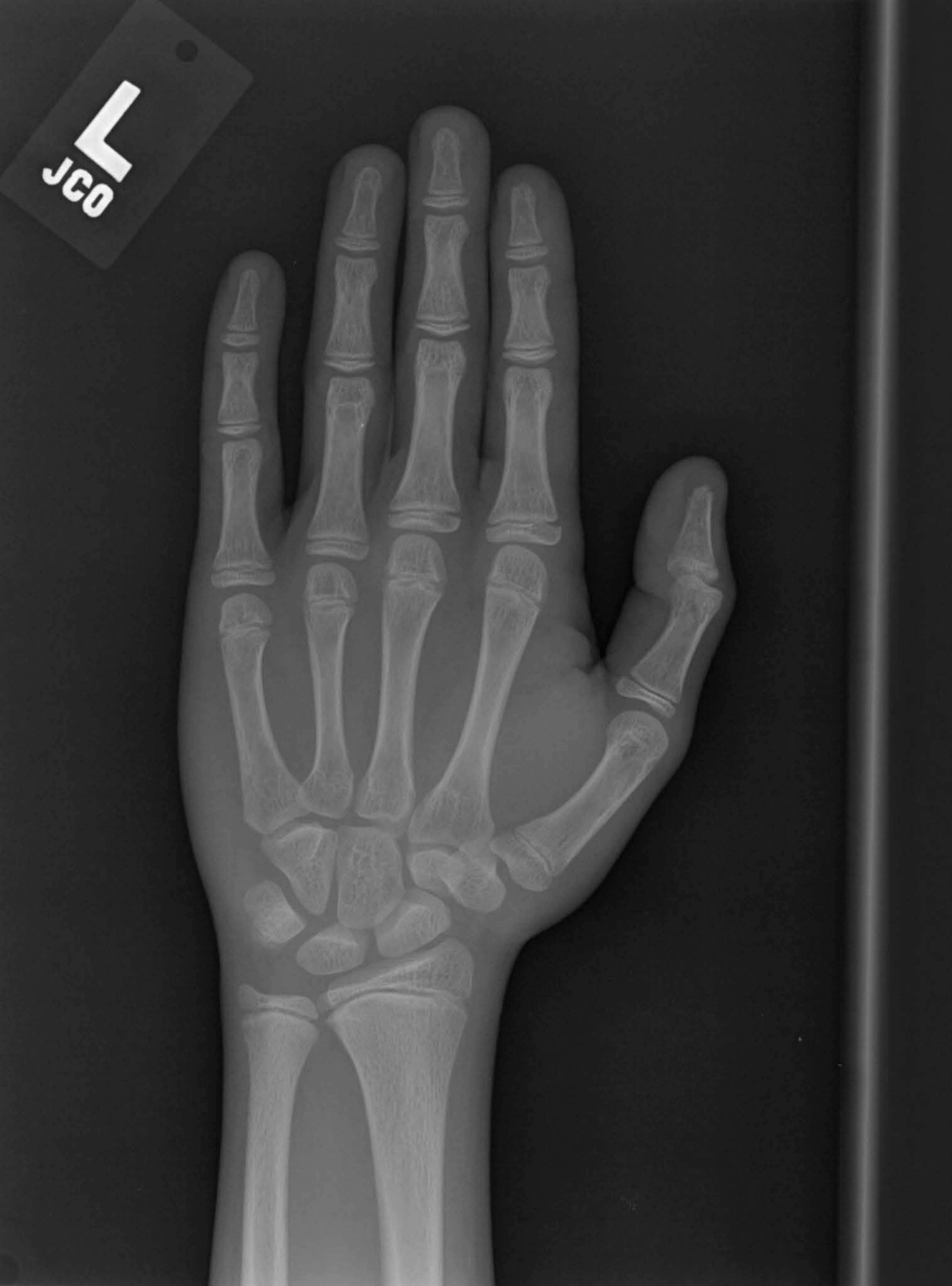}&
\includegraphics[width=1.1in,height=1.2in]{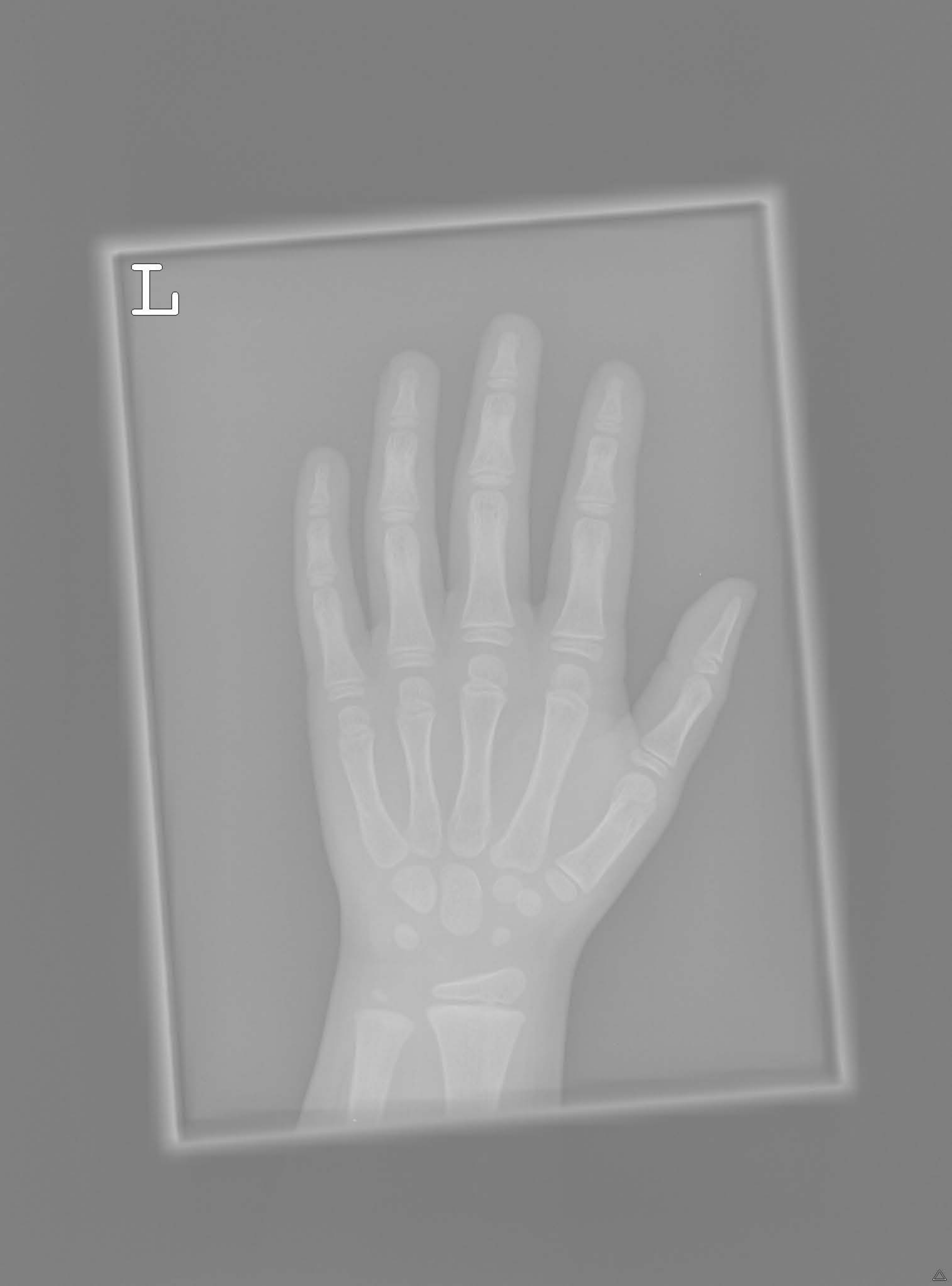}&
\includegraphics[width=1.1in,height=1.2in]{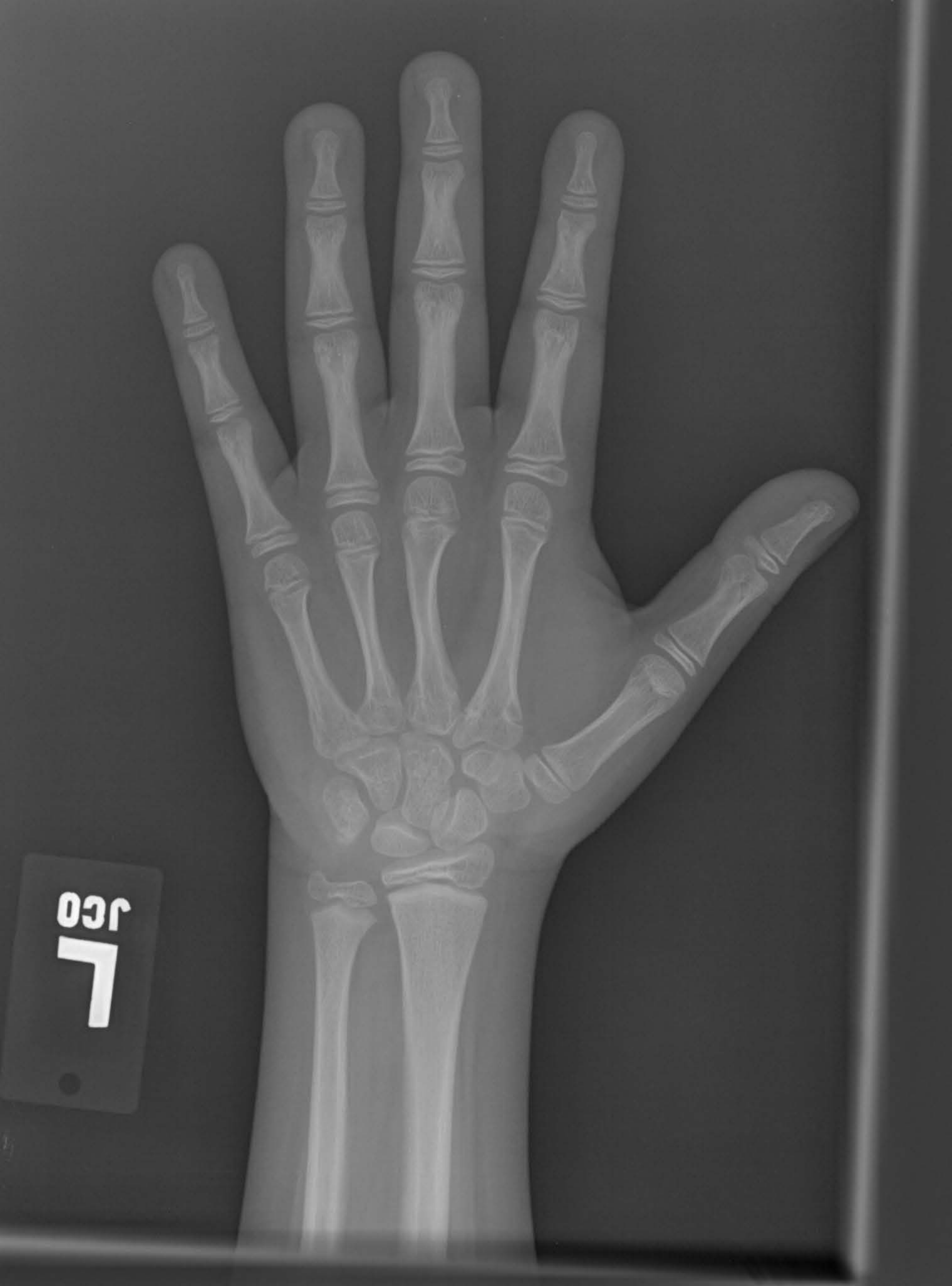}\\
\includegraphics[width=1.1in,height=1.2in]{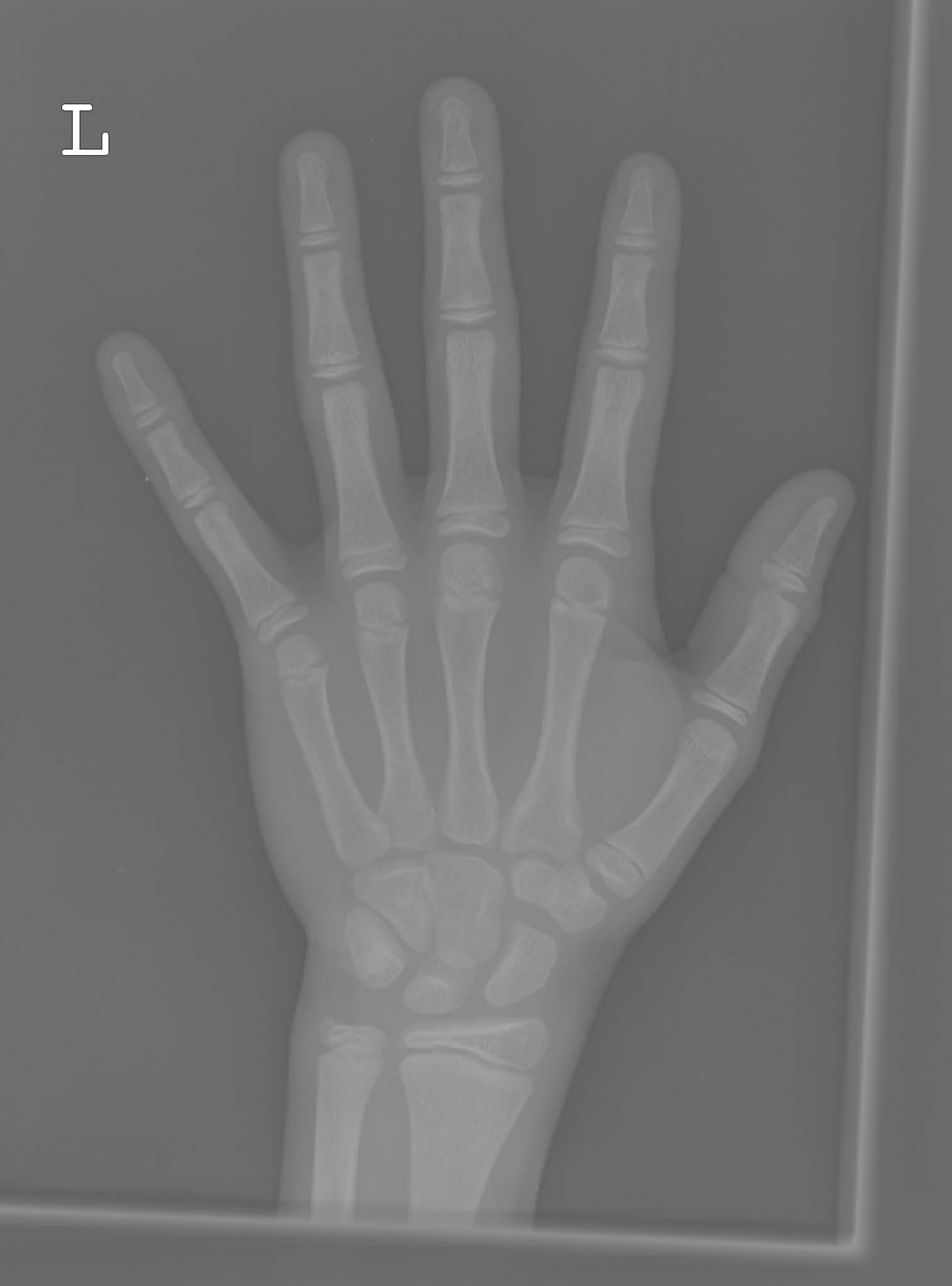}&
\includegraphics[width=1.1in,height=1.2in]{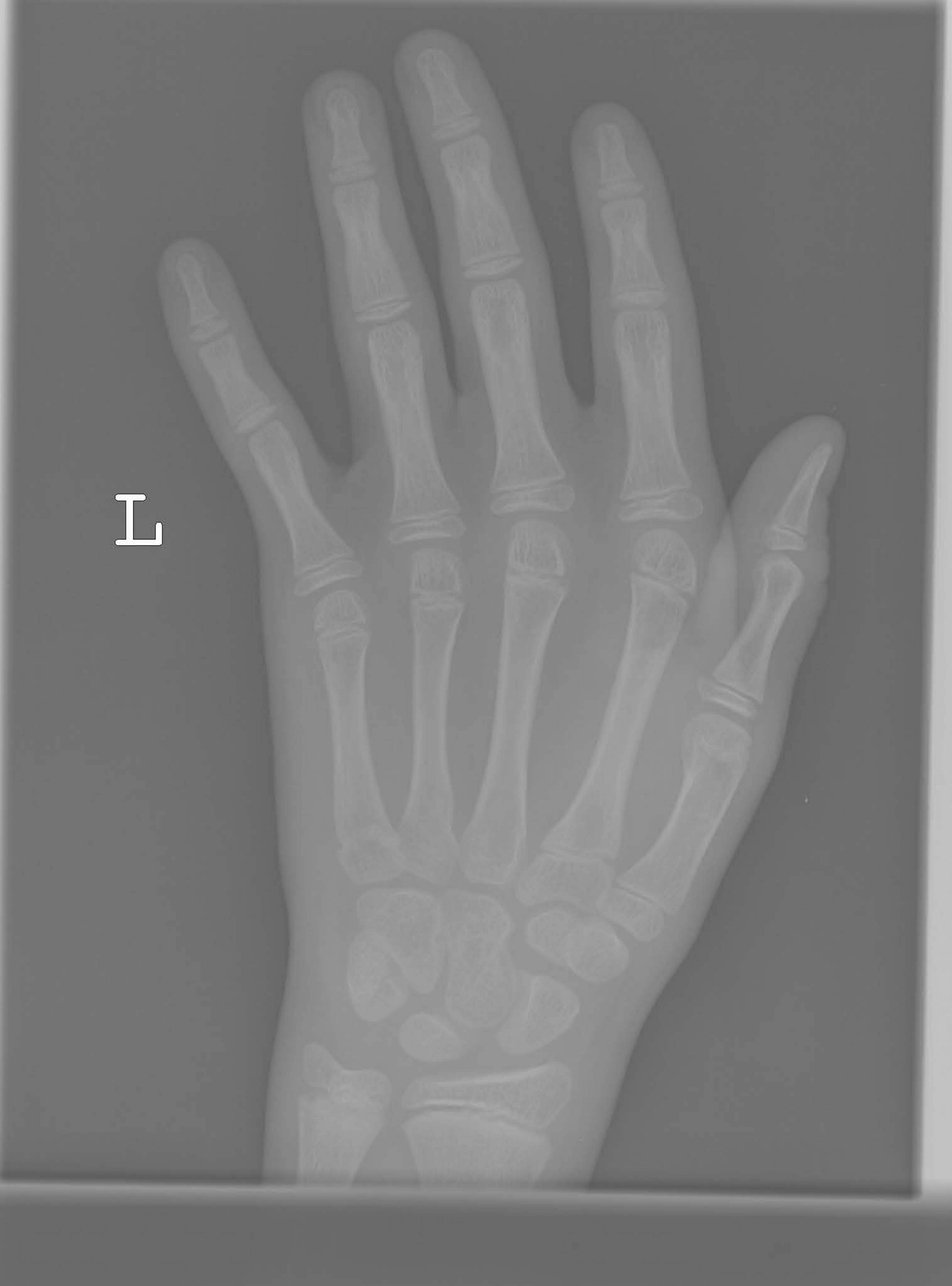}&
\includegraphics[width=1.1in,height=1.2in]{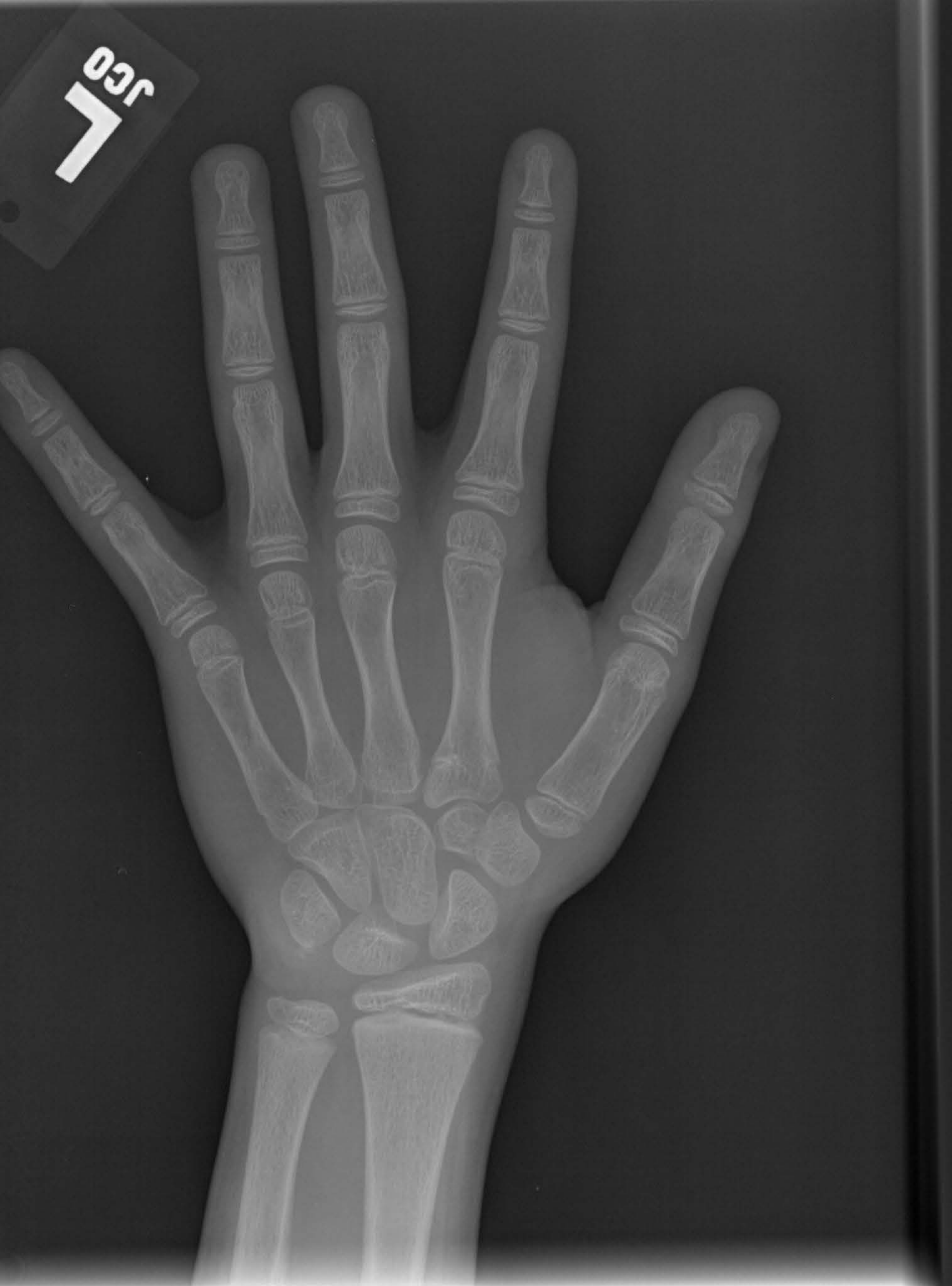}
\end{tabular}
\caption{Sample radiograph images from the RSNA dataset.}
\label{Fig:Examples_X-ray_images}
\end{figure}

\begin{figure}[!htb]
\centering
\includegraphics[width=3.5in,height=2.6in]{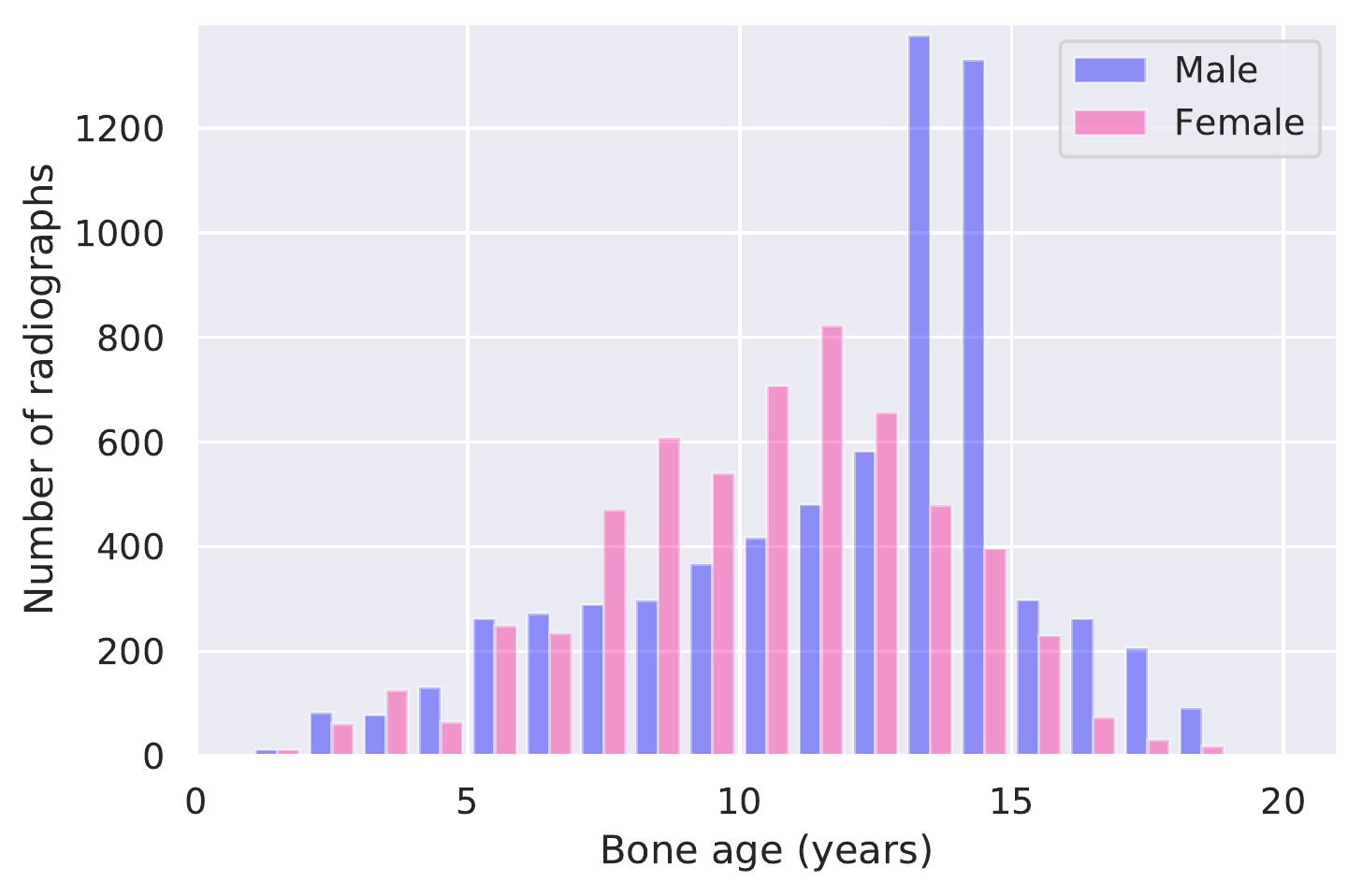}
\caption{Bone age distributions of male and female patients in the training set.}
\label{Fig:AgeDistribution}
\end{figure}

\medskip
\noindent{\textbf{Dataset:}}\quad The effectiveness of RidgeNet is evaluated on the Radiological Society of North America (RSNA) bone age dataset~\cite{Halabi:19}. RSNA consists of 14,236 hand radiographs of both male and female patients. The radiographs were acquired from Stanford Children's and Colorado Children's Hospitals at different times and under different conditions, and there are 12,611 images in the training set, 1,425 images in the validation set, and 200 images in the test set. Sample hand radiographs from the RSNA dataset are shown in Figure~\ref{Fig:Examples_X-ray_images}. As can be seen, the images were acquired at different times and under different conditions with varying size, background, contrast, brightness, and hand orientation. The training set contains 5778 radiographs of female patients and 6833 radiographs of male patients. As shown in Figure~\ref{Fig:AgeDistribution}, the bone ages for male and female patients are not uniformly distributed.

\medskip
\noindent{\textbf{Implementation details:}}\quad All the experiments were performed on Google Colaboratory, a free Jupyter notebook environment with 2 CPUs Intel(R) Xeon(R) 2.2GHz, 13GB RAM, and an NVIDIA Tesla K80 GPU accelerator. The algorithms were implemented in Python using Keras with Tensorflow backend. We used data augmentation to improve network accuracy and avoid overfitting by expanding the size of a training dataset. This is usually done by creating modified versions of the input images in the dataset through random transformations, including horizontal and vertical flip, brightness and zoom augmentation, horizontal and vertical shift augmentation, and rotation. For image annotation, we chose 80 images with different scales and orientations, and we used a polygon region shape to annotate the hand in each radiograph. Since we used the COCO dataset format for annotation, the annotated images are saved in JavaScript Object Notation (JSON) format for our custom dataset of segmented hand radiographs.

For feature extraction, we use the pre-trained VGG-19 convolutional neural network with input image size $512\times 512\times 3$, and rectified linear unit (ReLU) as an activation function. For ridge regression, the regularization parameter is $\lambda = 10^{-4}$, which was obtained via cross-validation. The proposed regression network was trained for 160 epochs using Adam optimizer with initial learning rate of $10^{-4}$ and batch size of 32, and using the mean square error as a loss function. The learning rate was multiplied by factor of 0.8 automatically when validation loss plateaued for 3 epochs. The value of MSE is computed as the average of the squared differences between the predicted and actual values.

\begin{figure}[!htb]
\centering
\includegraphics[scale=0.67]{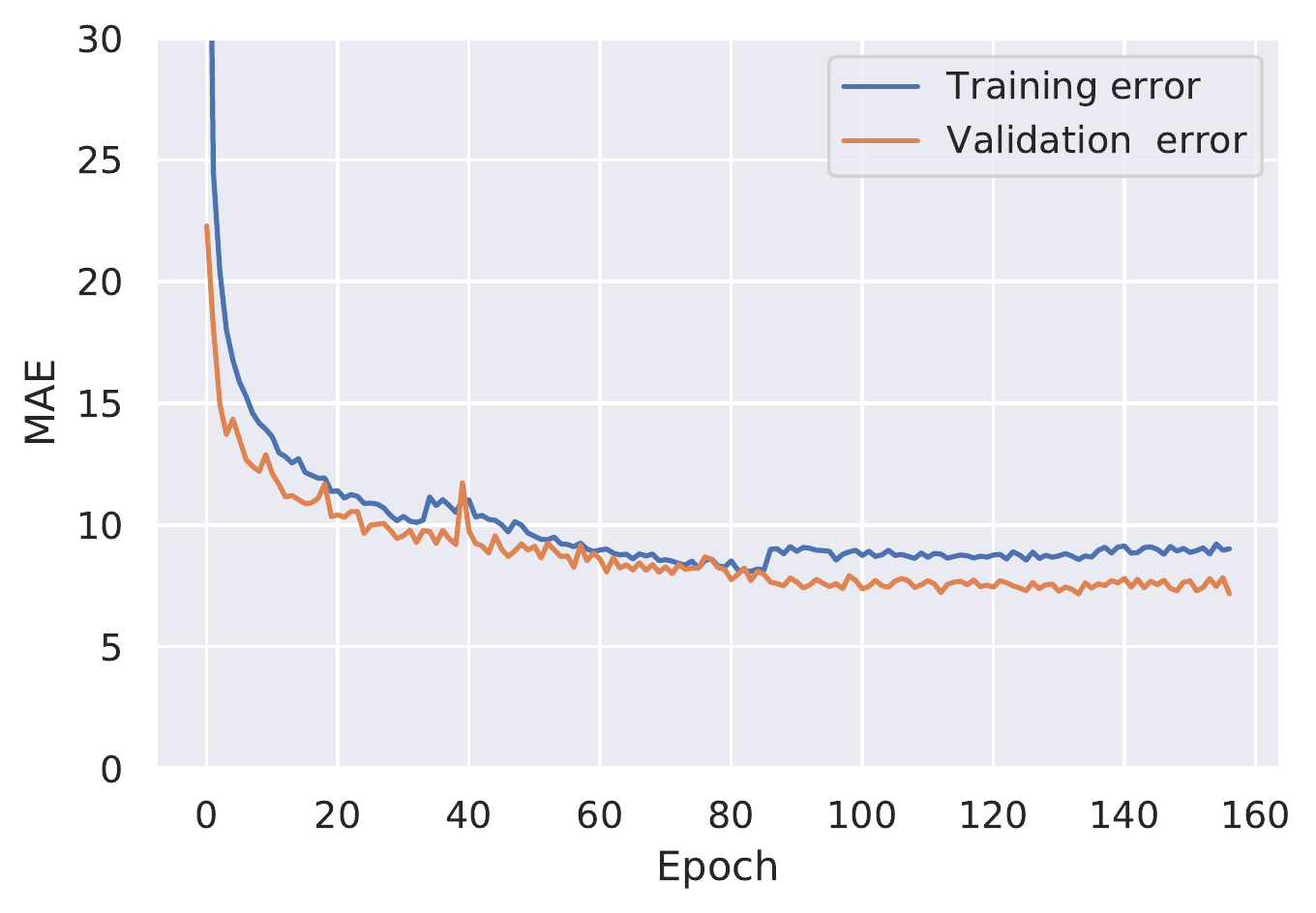}
\caption{Model training history.}
\label{Fig:TrainingErros}
\end{figure}

The MSE values are recorded at the end of each epoch on the training and/or validation sets. The performance of the RidgeNet model over training epochs on the training and validation sets is evaluated using the mean absolute error (MAE) metric, as shown in Figure~\ref{Fig:TrainingErros}. As can be seen, the RideNet model has comparable performance on both training and validation sets, indicating the higher predictive accuracy of the proposed model.

\medskip
\noindent{\textbf{Performance evaluation metrics:}}\quad The performance of a regression model is usually assessed by applying it to test data with known target values and comparing the predicted values with the known values. To this end, we use mean absolute error (MAE), root mean square error (RMSE), and root mean squared percentage error (RMSPE) as evaluation metrics, which are defined as follows:
\begin{equation}
\text{MAE} = \frac{1}{N}\sum_{i=1}^{N}|y_i - \hat{y}_i|
\label{Eq:MAE}
\end{equation}

\begin{equation}
\text{RMSE}= \sqrt{\frac{1}{N}\sum_{i=1}^{N}(y_i - \hat{y}_i)^2}
\label{Eq:RMSE}
\end{equation}

\begin{equation}
\text{RMSPE}= \sqrt{\frac{1}{N}\sum_{i=1}^{N}\Big(\frac{y_i - \hat{y}_i}{y_i}\Big)^2}
\label{Eq:RMSE}
\end{equation}
where $N$ is the number of samples in the test set, $y_i$ is the actual (ground truth) value, and $\hat{y}_i$ is the model's predicted value. A small value of an evaluation metrics indicates a better performance of the model.

\medskip
\noindent{\textbf{Baseline methods:}}\quad We compared the effectiveness of the proposed model with several deep learning based approaches, including a pre-trained convolutional neural network with Gaussian process regression~\cite{van:18}, an instance segmentation model with residual attention network~\cite{Wu:19}, regression/classification models~\cite{iglovikov:18}, and U-Net based model~\cite{pan:20}. The regression model presented in~\cite{iglovikov:18} is similar to the classification model, expect for the lat two layers and also the fact that a bone age is assigned a class. The penultimate layer is a softmax layer that returns a vector of probabilities for all the classes. In the last layer, these probabilities are multiplied by a vector of bone ages uniformly distributed over an interval of length equal to the total number of classes. The U-Net based model~\cite{pan:20} consists of image segmentation, feature extraction, and ensemble modules.

\subsection{Results}
\begin{figure}[!htb]
\centering
\includegraphics[scale=0.26]{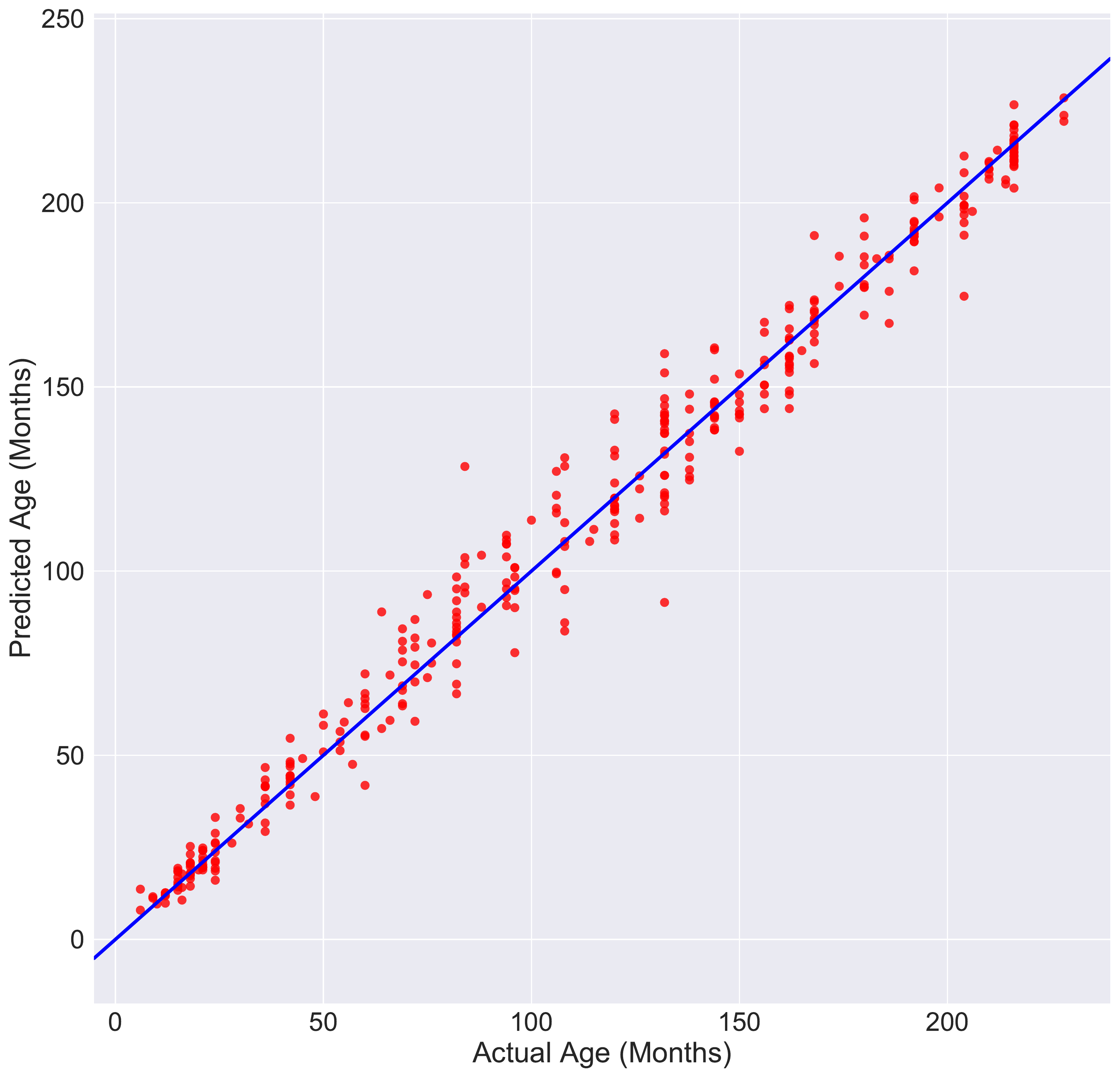}
\caption{Predicted bone age vs. actual bone age for both genders on the RSNA test set.}
\label{Fig:Scatter_mixed}
\end{figure}

\begin{figure}[!htb]
\centering
\includegraphics[scale=0.26]{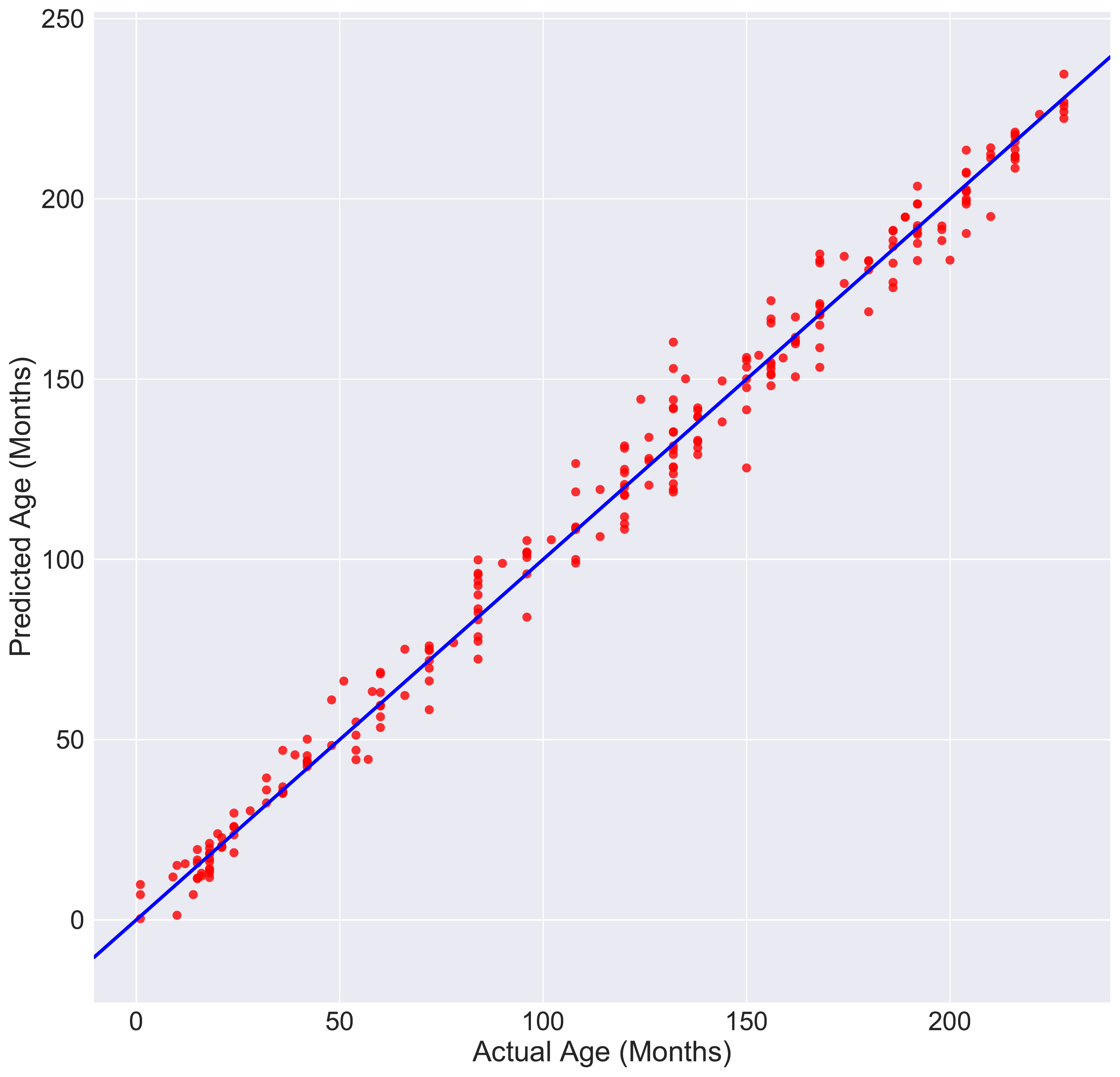}
\caption{Predicted bone age vs. actual bone age for male patients on the RSNA test set.}
\label{Fig:Scatter_Males}
\end{figure}

\begin{figure}[!htb]
\centering
\includegraphics[scale=0.26]{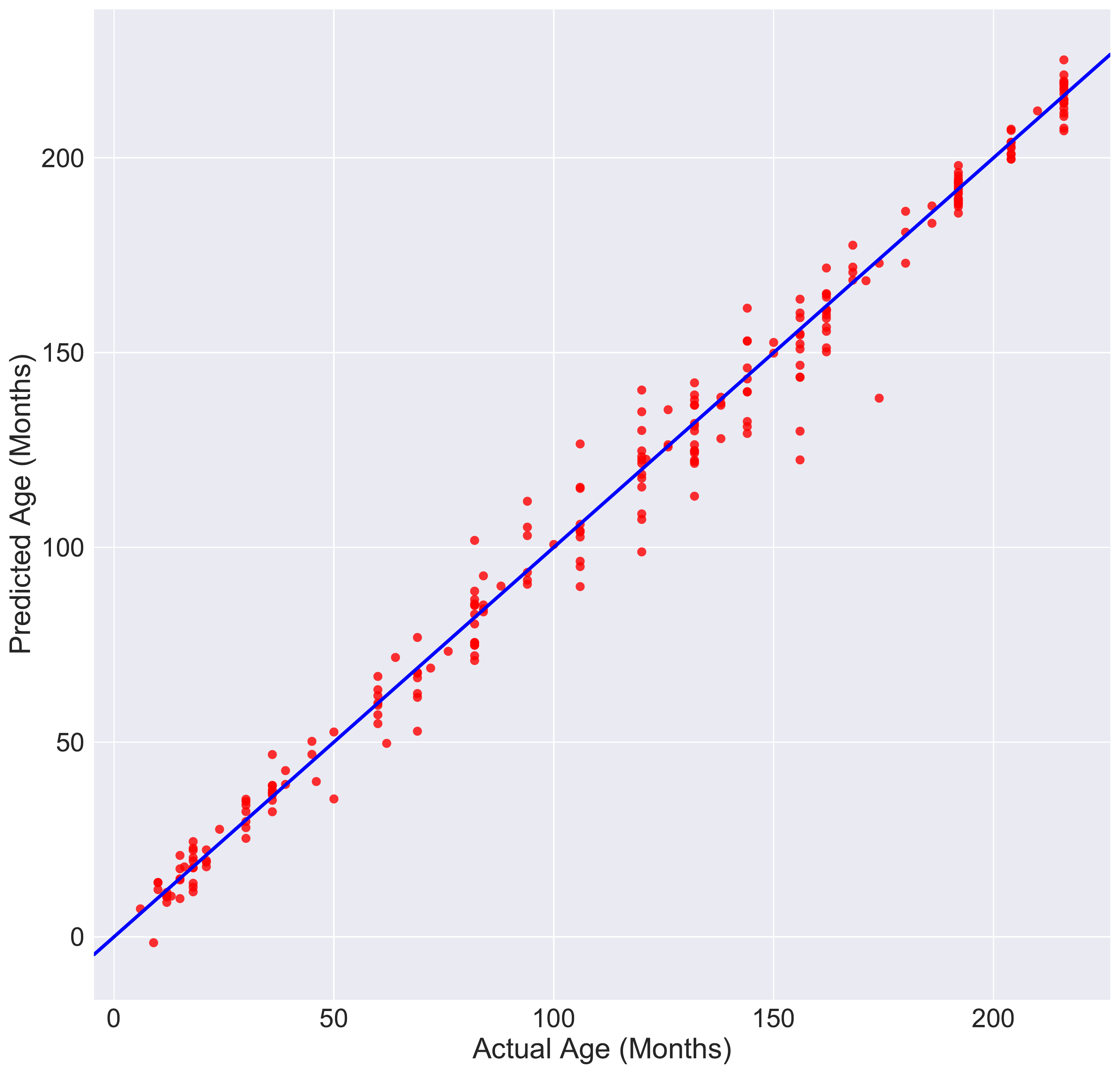}
\caption{Predicted bone age vs. actual bone age for female patients on the RSNA test set.}
\label{Fig:Scatter_Females}
\end{figure}

In this subsection, we report the prediction results obtained by our proposed approach on the RSNA bone age test set, and provide a comparison analysis with baseline methods. Figures~\ref{Fig:Scatter_mixed}-\ref{Fig:Scatter_Females} show the predicted bone age versus the actual bone age for both genders, males, and females, respectively, on the test set. The solid line depicts the actual bone age, while the dot points represent the predicted values. As can be seen, the predicted values align pretty well along the solid line, indicating a strong agreement between the actual bone age and predicted one. It is important to point out that the plot for both genders in Figure~\ref{Fig:Scatter_mixed} shows a noticeable variation in skeletal age prediction between 84 and 156 months. This variation is due largely to the fact that girls mature faster than boys during the early and mid-puberty stage (7- to 13-year old girls compared with 9- to 14-year old boys), and hence skeletal maturation changes occur earlier in females than in males. Notice also that there is little variation in skeletal age prediction during infancy (birth to 10-month old girls and birth to 14-month old boys), toddlers (10-month to 2-year old girls and 10-month to 3-year old boys), pre-puberty (2- to 7-year old girls and 3- to 9-year old boys) compared to late puberty (13- to 15-year old girls and 14- to 16-year old boys) and post-puberty (14- to 17-year old girls and 17- to 19-year old boys).

\begin{figure}[!htb]
\centering
\includegraphics[scale=0.45]{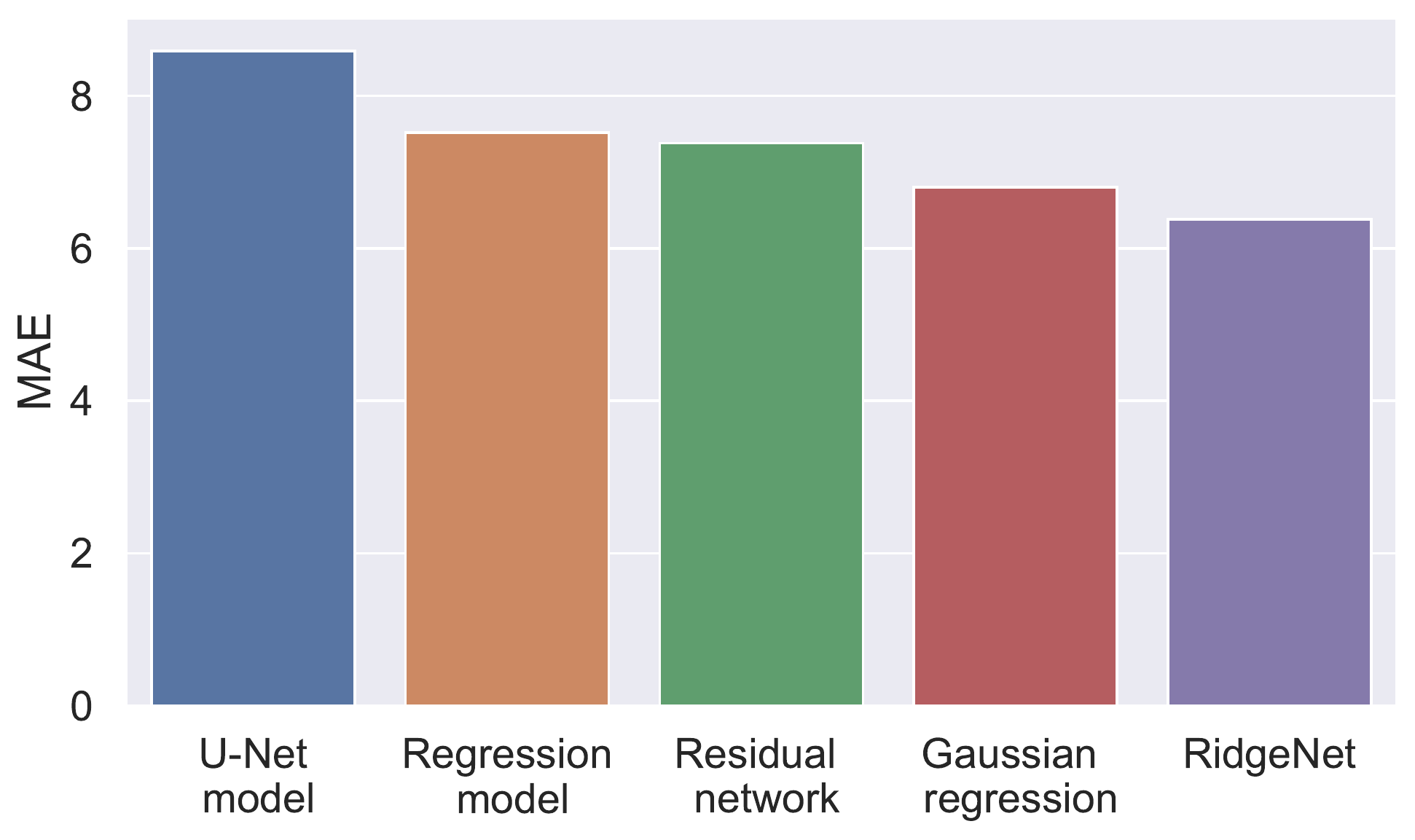}
\caption{MAE results for RidgeNet and baseline methods for both genders on the RSNA test set.}
\label{Fig:MAE_mixed}
\end{figure}

\begin{figure}[!htb]
\centering
\includegraphics[scale=0.45]{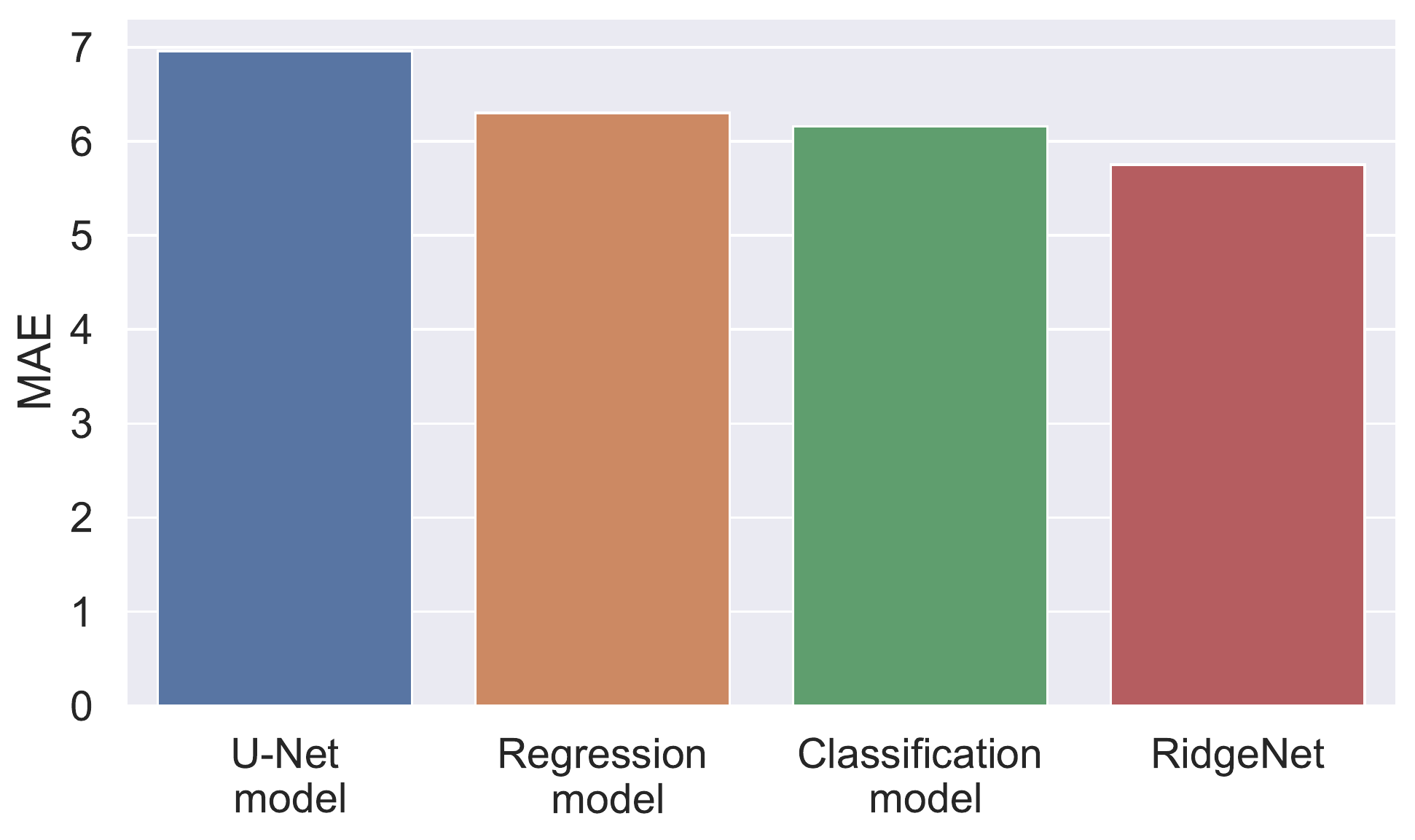}
\caption{MAE results for RidgeNet and baseline methods for male patients on the RSNA test set.}
\label{Fig:MAE_males}
\end{figure}

\begin{figure}[!htb]
\centering
\includegraphics[scale=0.45]{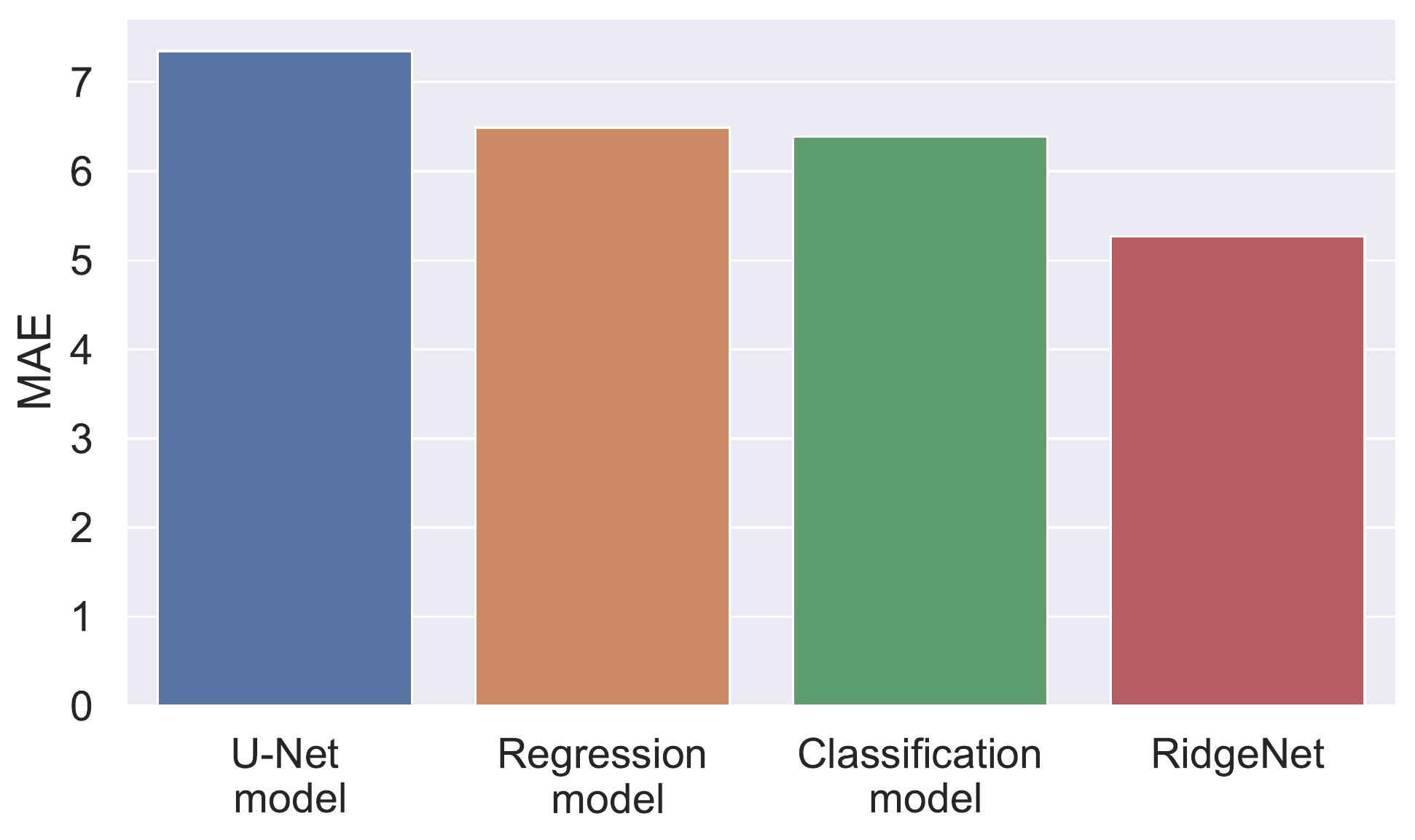}
\caption{MAE results for RidgeNet and baseline methods for female patients on the RSNA test set.}
\label{Fig:MAE_females}
\end{figure}

The bar plots of the MAE values obtained by RidgeNet and the baseline methods for both genders are displayed in Figure~\ref{Fig:MAE_mixed}, which shows that the proposed approach performs the best, yielding the lowest MAE of 6.38 months. Similarly, the bar plots shown in Figures~\ref{Fig:MAE_males} and~\ref{Fig:MAE_females} for male and female patients, respectively, indicate that RidgeNet substantially outperforms both the regression and classification models. As can be seen in these figures, the proposed approach yields the best overall results. In addition, it is worth pointing out that gender-specific regression by RidgeNet yields lower MAE values, with a MAE of 5.27 months for females and a MAE of 3.75 months for males compared to a MAE of 6.38 months for both sexes.

The evaluation results using the RMSE and RMSPE metrics are shown in Table~\ref{Table:RMSE_and_RMSPE}. As can be seen, RidgNet yields the lowest RMSPE value in the case of both genders.
\begin{table}[!htb]
\centering\small
\caption{Evaluation results for RidgeNet on the RSNA test set using the RMSE and RMSPE metrics.}
\medskip
\label{Table:RMSE_and_RMSPE}
\begin{tabular}{lcc}
\toprule
Gender  & RMSE   & RMSPE  \\ \hline
Both   & 8.70   & 2.71  \\
Males   & 7.00   & 3.06 \\
Females & 7.49      & 3.08   \\

\bottomrule
\end{tabular}
\end{table}

\begin{figure}[!htb]
\setlength{\tabcolsep}{.8em}
\centering
\begin{tabular}{cccc}
\includegraphics[width=0.8in,height=1.1in]{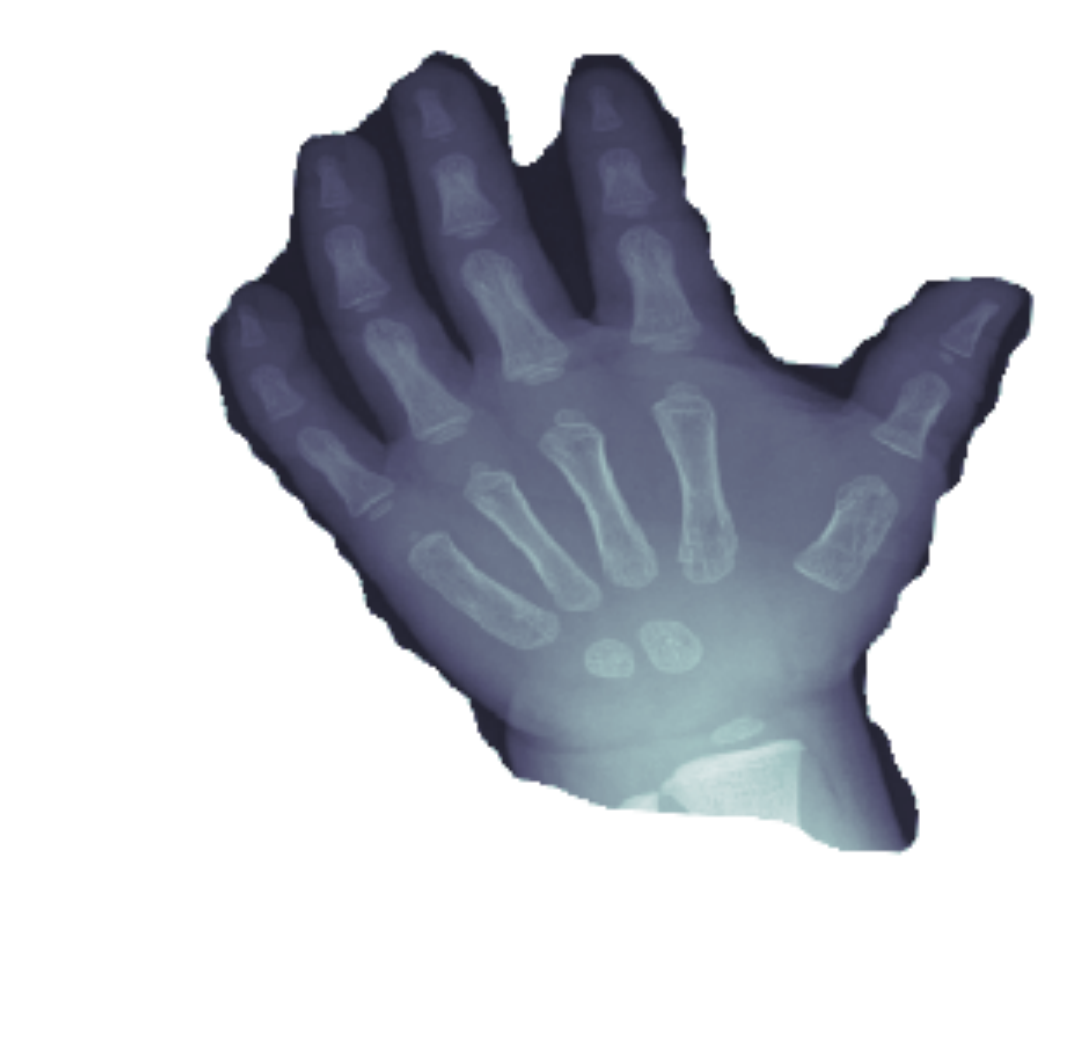}&
\includegraphics[width=0.8in,height=1.1in]{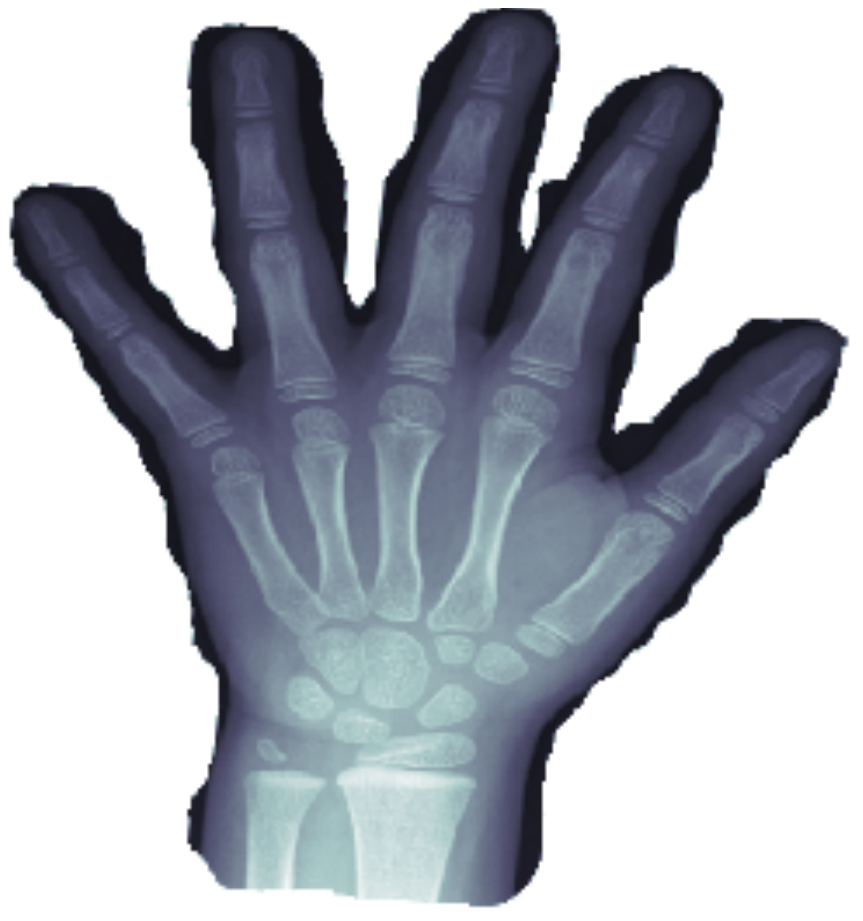}&
\includegraphics[width=0.8in,height=1.1in]{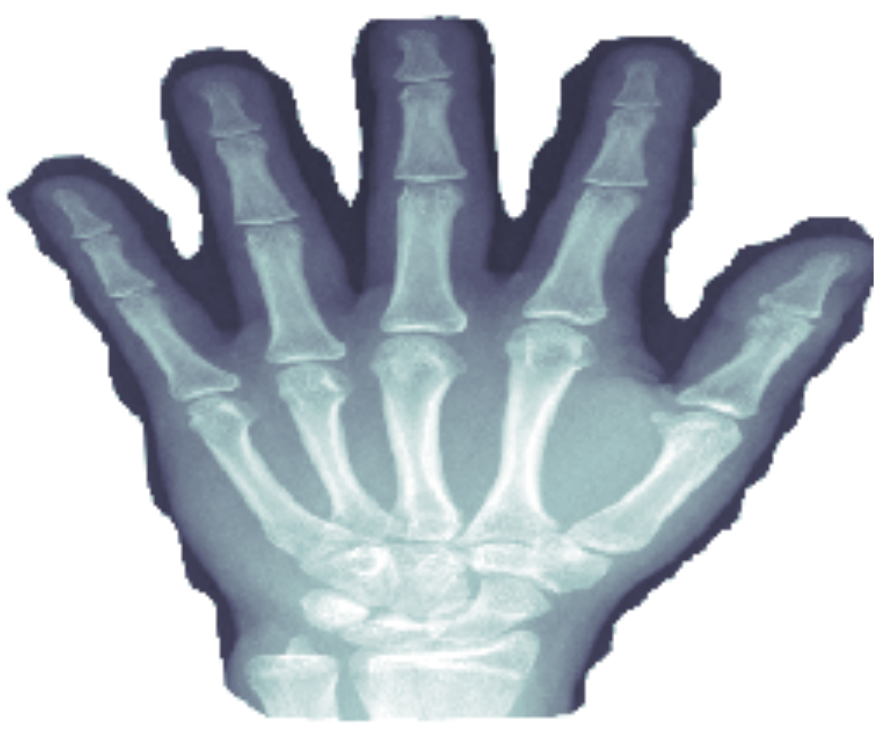}&\\
\scriptsize{Actual age: 82} & \scriptsize{Actual age: 18} & \scriptsize{Actual age: 168} \\
\scriptsize{Predicted age: 85.9} & \scriptsize{Predicted age: 20.1}  & \scriptsize{Predicted age: 167.5} \\\\
\includegraphics[width=0.8in,height=1.1in]{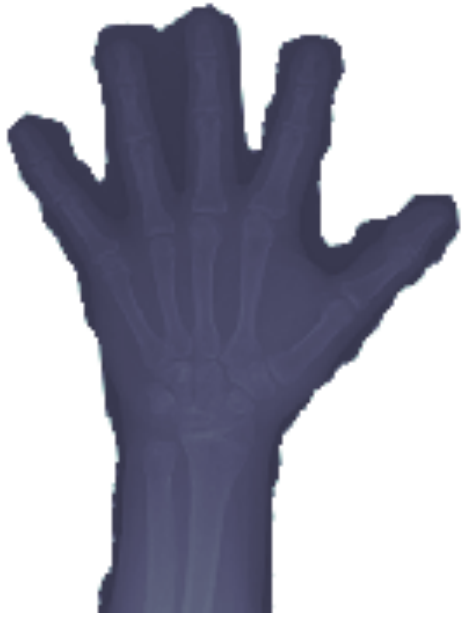}&
\includegraphics[width=0.8in,height=1.1in]{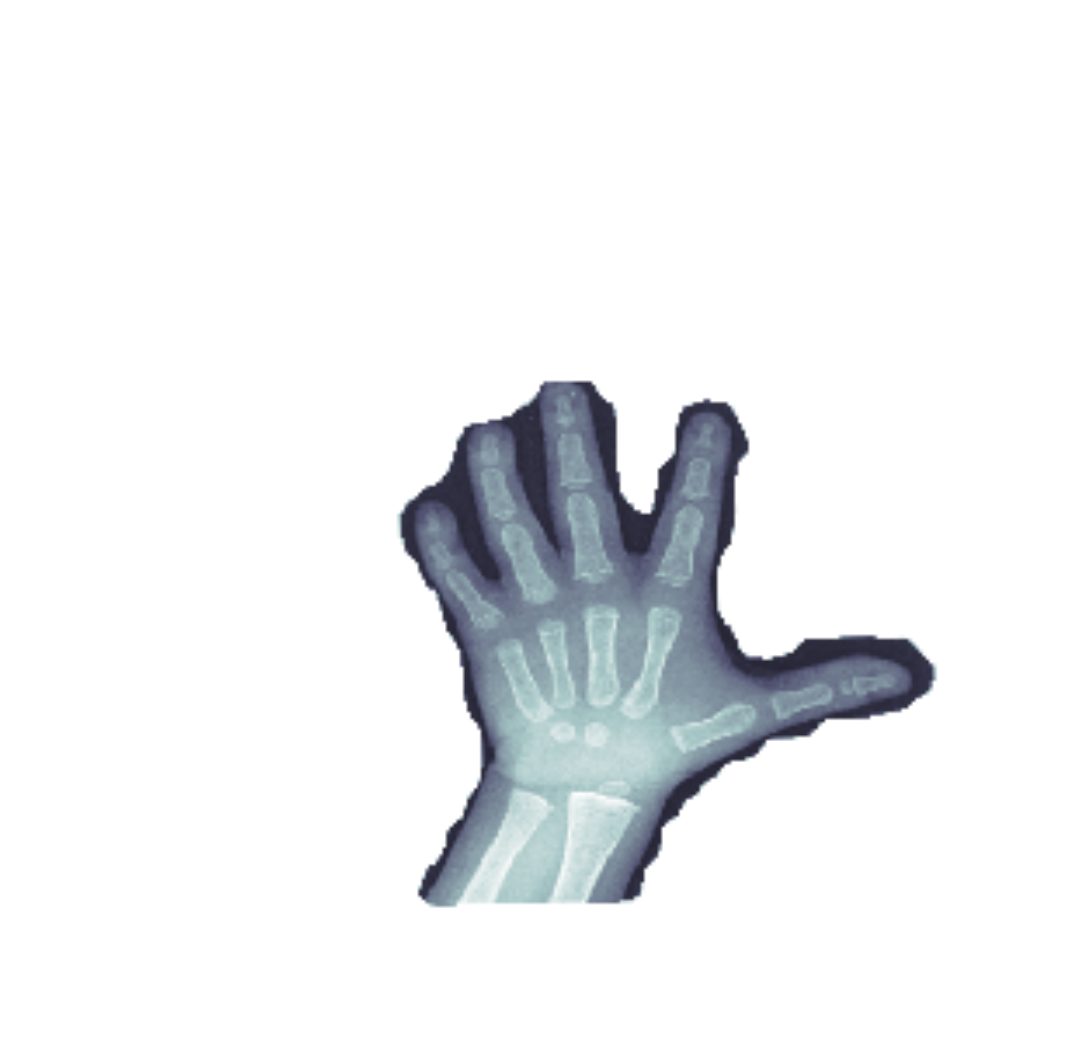}&
\includegraphics[width=0.8in,height=1.1in]{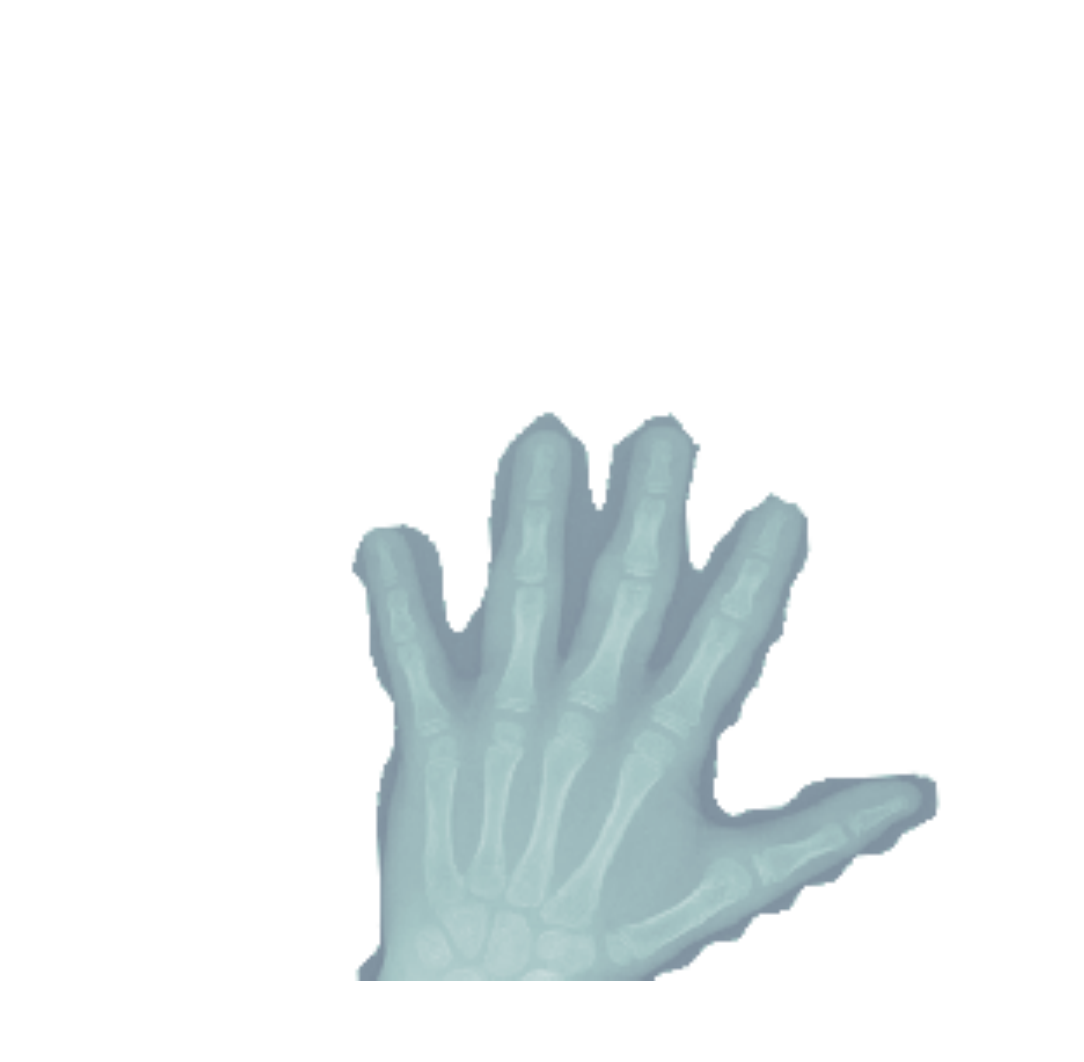}&\\
\scriptsize{Actual age: 192} & \scriptsize{Actual age: 34} & \scriptsize{Actual age: 88}  \\
\scriptsize{Predicted age: 190.5} & \scriptsize{Predicted age: 36.3}  & \scriptsize{Predicted age: 91.3}\\\\
\includegraphics[width=0.8in,height=1.1in]{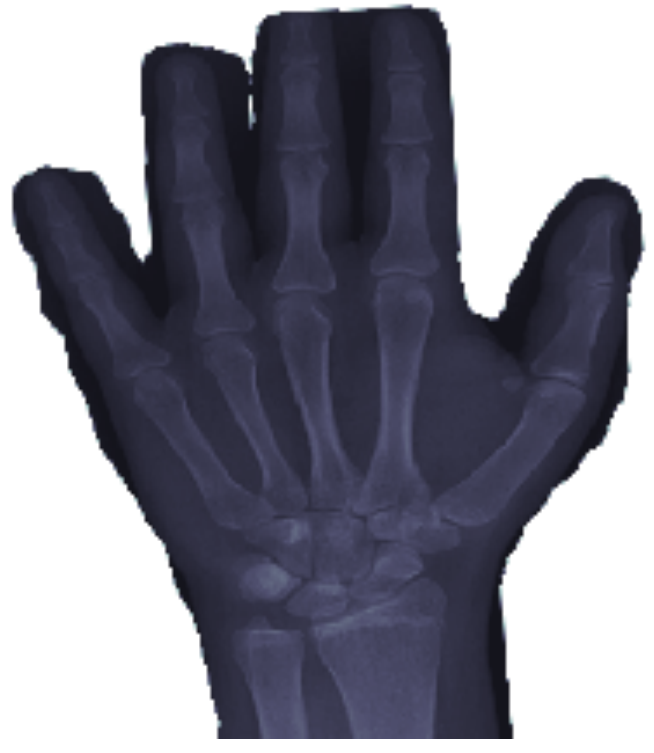}&
\includegraphics[width=0.8in,height=1.1in]{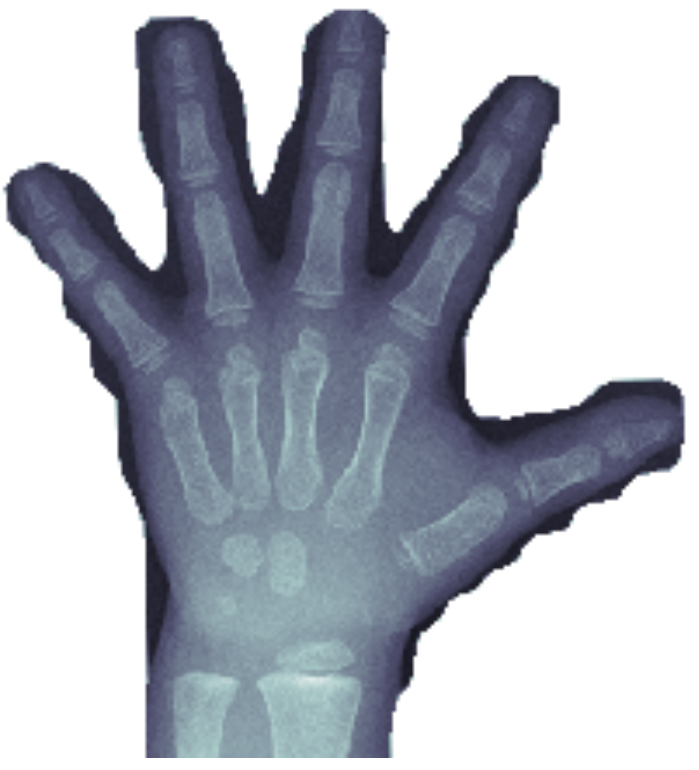}&
\includegraphics[width=0.8in,height=1.1in]{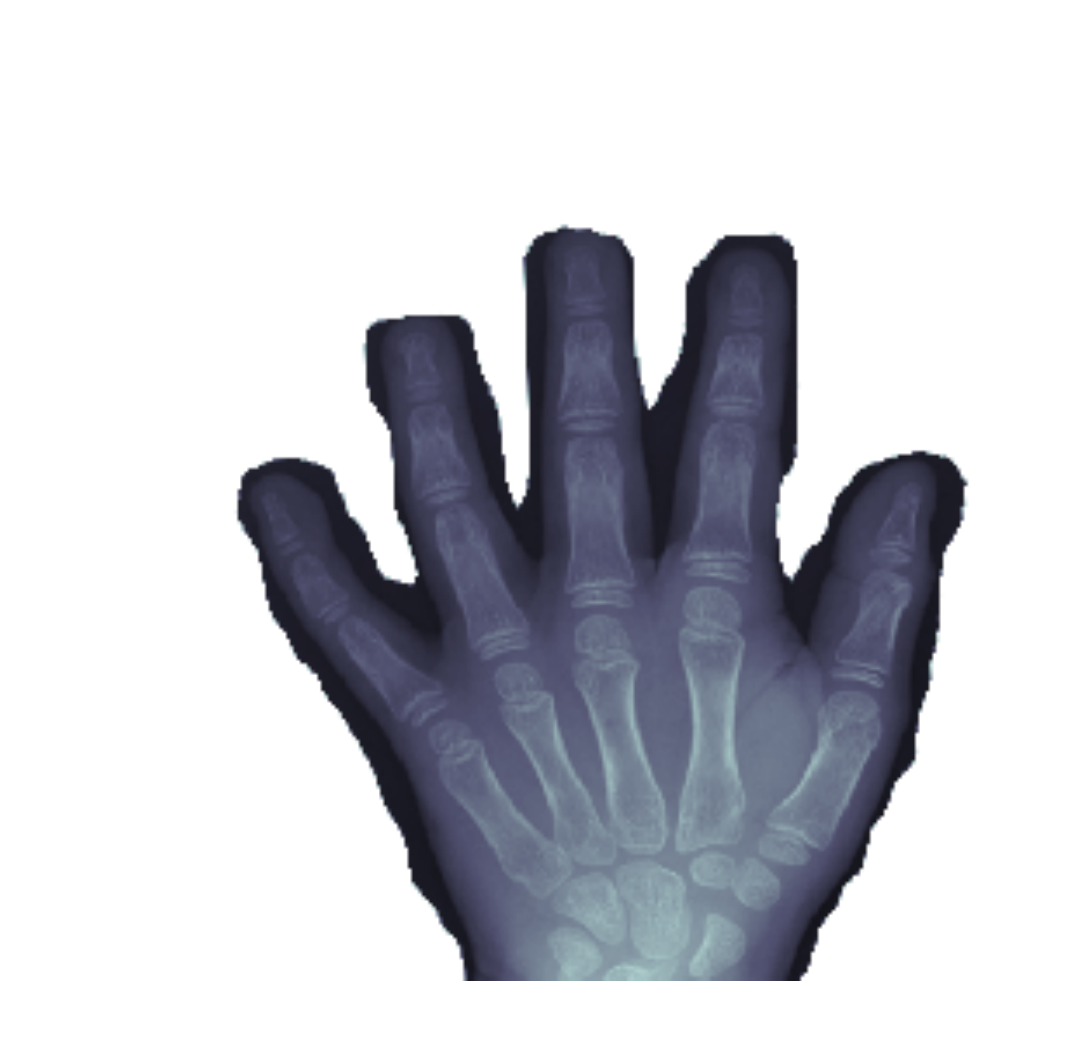}&\\
\scriptsize{Actual age: 192} & \scriptsize{Actual age: 30} & \scriptsize{Actual age: 82}  \\
\scriptsize{Predicted age: 194.5} & \scriptsize{Predicted age: 33.8}  & \scriptsize{Predicted age: 82.8}\\
\end{tabular}
\caption{Actual and predicted bone age of sample images from the test RSNA dataset. The first row displays images of both genders, while the second and the third row show the images of males and females, respectively}
\label{Fig:Acutal_Predicted_Image}
\end{figure}

Figure~\ref{Fig:Acutal_Predicted_Image} shows the actual and predicted bone ages of some segmented images from the test set. As can be seen, the predicted bone ages are quite comparable to the actual bone ages, indicating the effectiveness of the proposed RidgeNet model in bone age assessment.

\subsection{Feature Visualization and Analysis}
Understanding and interpreting the predictions made by a deep learning model provides valuable insights into the input data and the features learned by the model so that the results can be easily understood by human experts. To visually explain the predictions obtained by the proposed RidgeNet model, we apply Smooth Grad-CAM++, an enhanced gradient weighted class activation mapping~\cite{Omeiza:19} that highlights the most influential features affecting the predictions of the model on various radiographs from different skeletal maturation stages, including pre-puberty, early and mid-puberty, late puberty, and post-puberty. Gradient weighted class activation mapping produces heat maps via a linear weighted combination of the activation maps to highlight discriminative image regions, indicating where the deep neural network bases its predictions. The Smooth Grad-CAM++ method combines concepts from Grad-CAM~\cite{Selvaraju:17}, SmoothGrad~\cite{Smilkov:17} and Grad-CAM++~\cite{Chattopadhay:18} to produce improved, visually sharp maps for specific layers, subset of feature maps and neurons of interest.

\begin{figure}[!htb]
\centering
\includegraphics[scale=0.78]{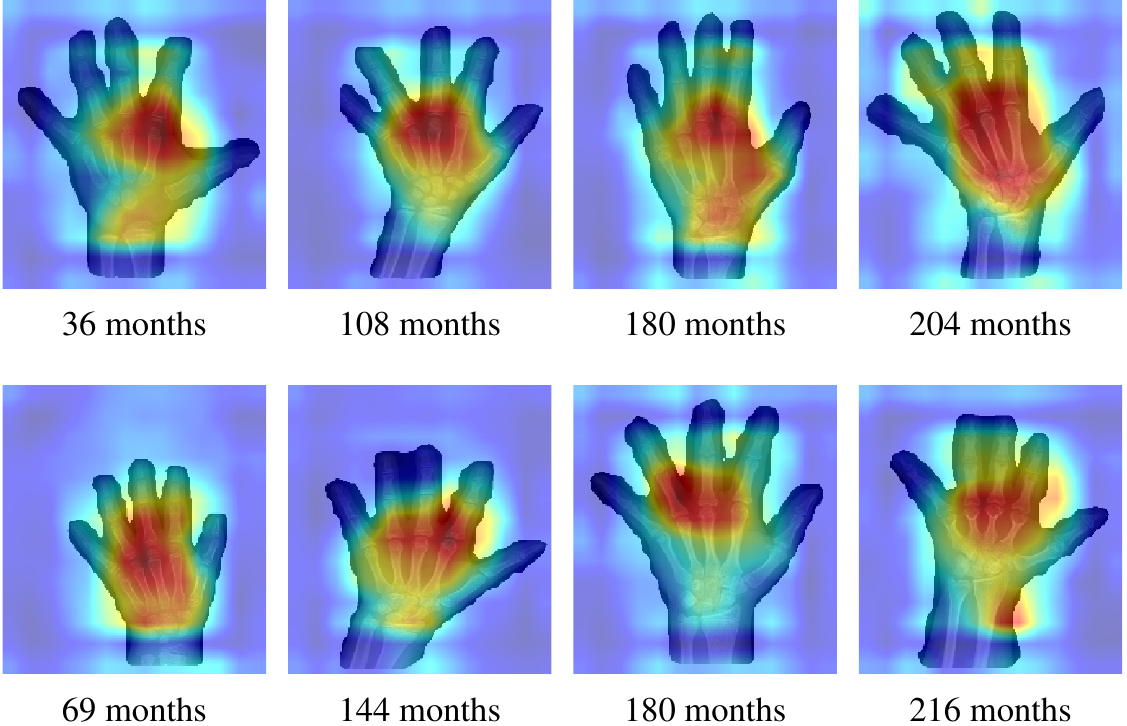}
\caption{Smooth Grad-CAM++ heat maps for female (top) and male (bottom) patients at the pre-puberty, early and mid-puberty, late puberty, and post-puberty stages.}
\label{Fig:HeatMaps}
\end{figure}

Figure~\ref{Fig:HeatMaps} shows the class activation maps for RidgeNet using Smooth Grad-CAM++ for female (top) and male (bottom) patients at the pre-puberty, early and mid-puberty, late puberty, and post-puberty stages. As can be seen, Smooth Grad-CAM++ highlights the regions of the radiographs that play a significant role in bone age assessment for different age categories. The red regions contribute the most to the predictions obtained by RidgeNet, and the deep blue color indicates the lowest contribution. In the early and mid-puberty stage, for instance, the class activation map emphasizes the metacarpal bones and proximal phalanges. In the post-puberty stage, the RidgeNet model focuses on carpal bones, meaning that they have the greatest impact on the regression results. These results are consistent with what radiologists consider the most suitable indicators of skeletal maturity during the different phases of postnatal development~\cite{Gilsanz:12}.

\subsection{Age Assessment on chest X-rays}
\begin{figure}[!htb]
\setlength{\tabcolsep}{.2em}
\centering
\begin{tabular}{ccc}
\includegraphics[width=1.1in,height=1.2in]{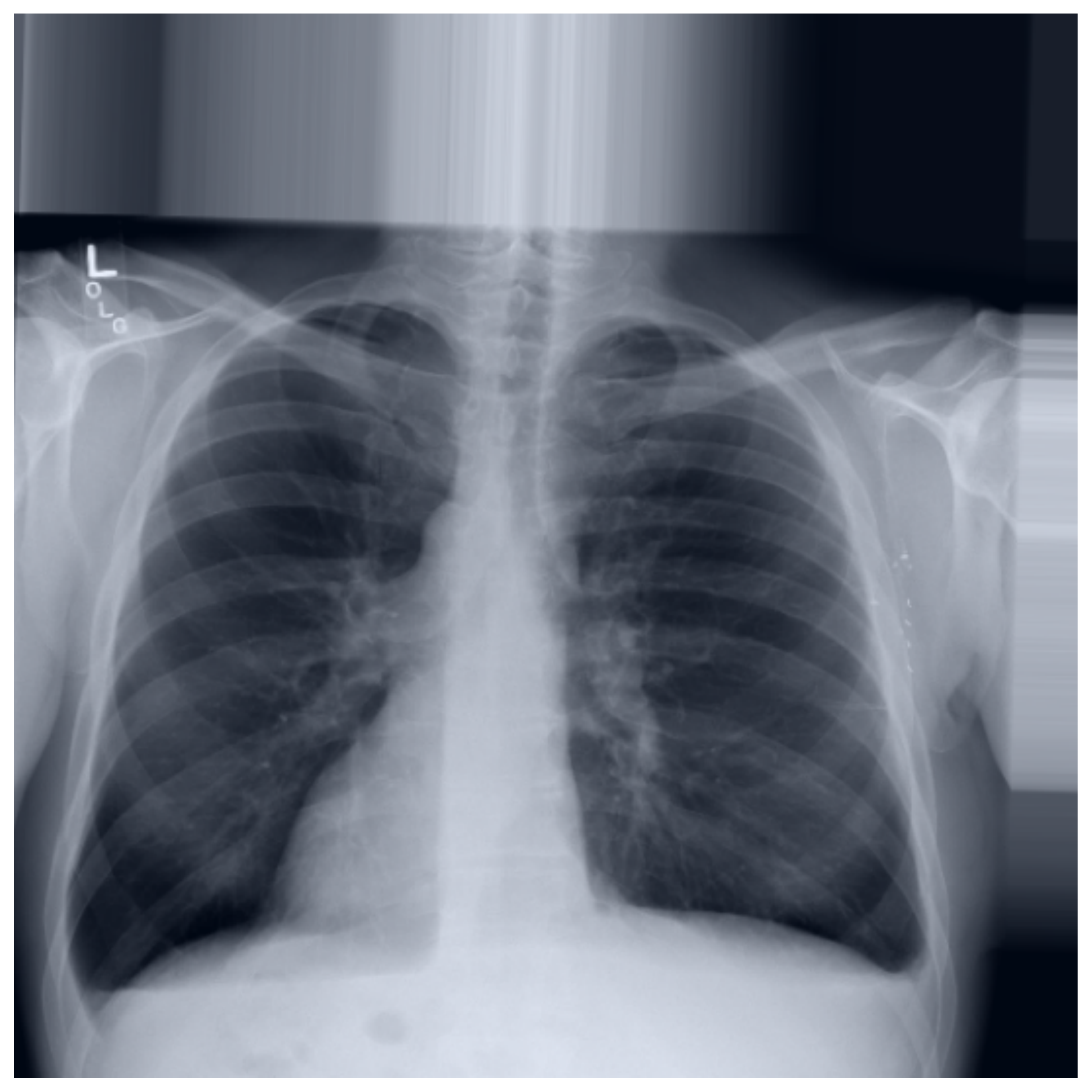}&
\includegraphics[width=1.1in,height=1.2in]{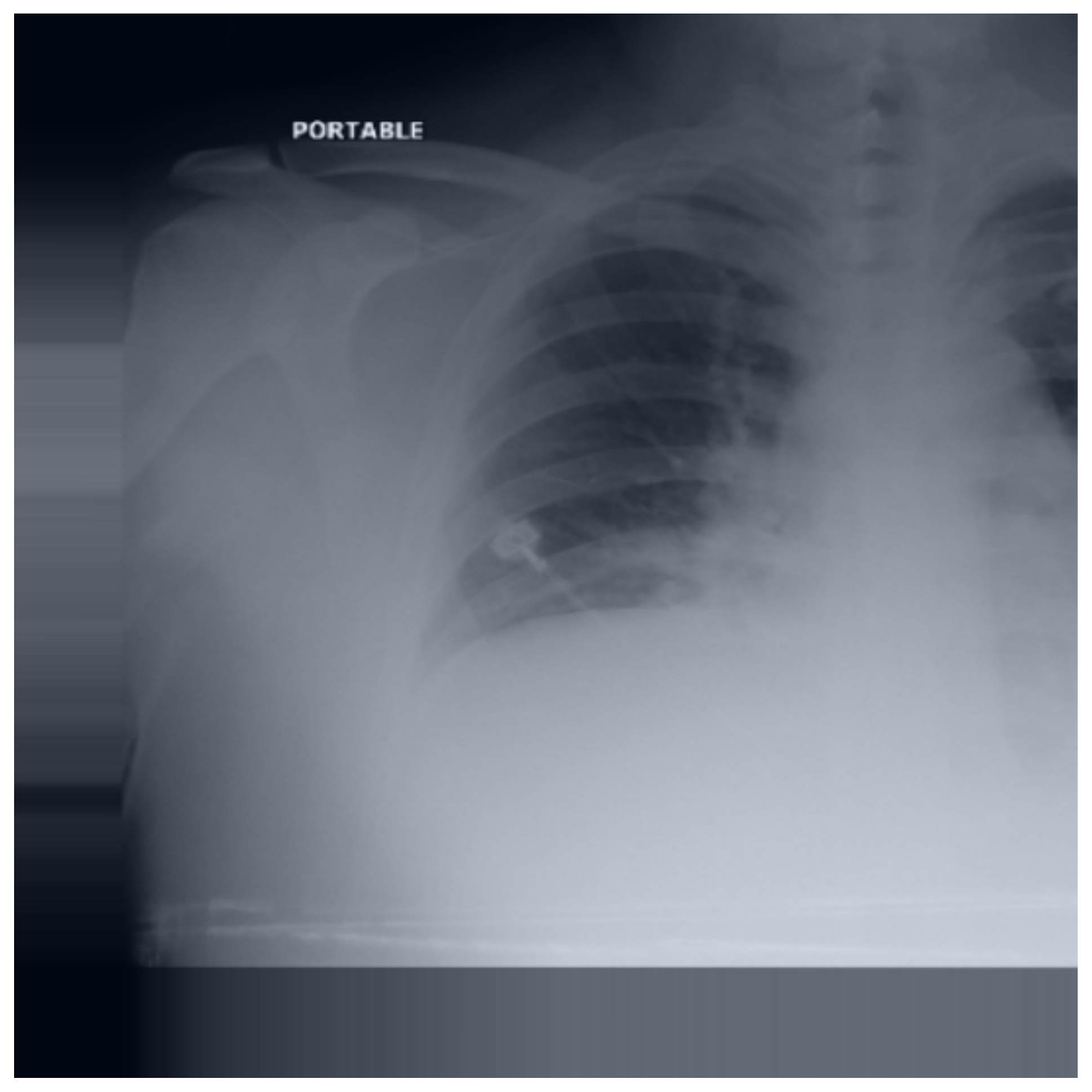}&
\includegraphics[width=1.1in,height=1.2in]{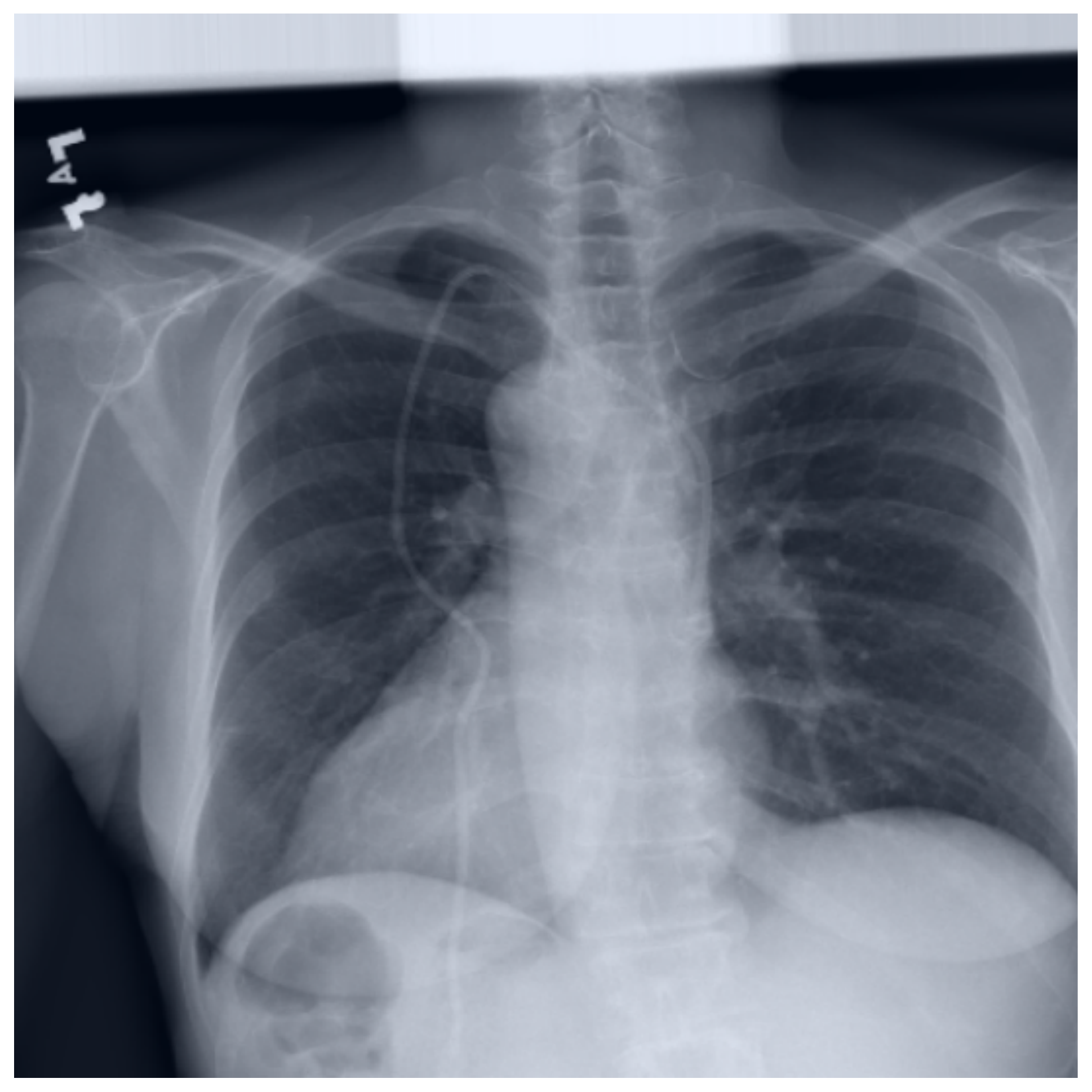}\\
\includegraphics[width=1.1in,height=1.2in]{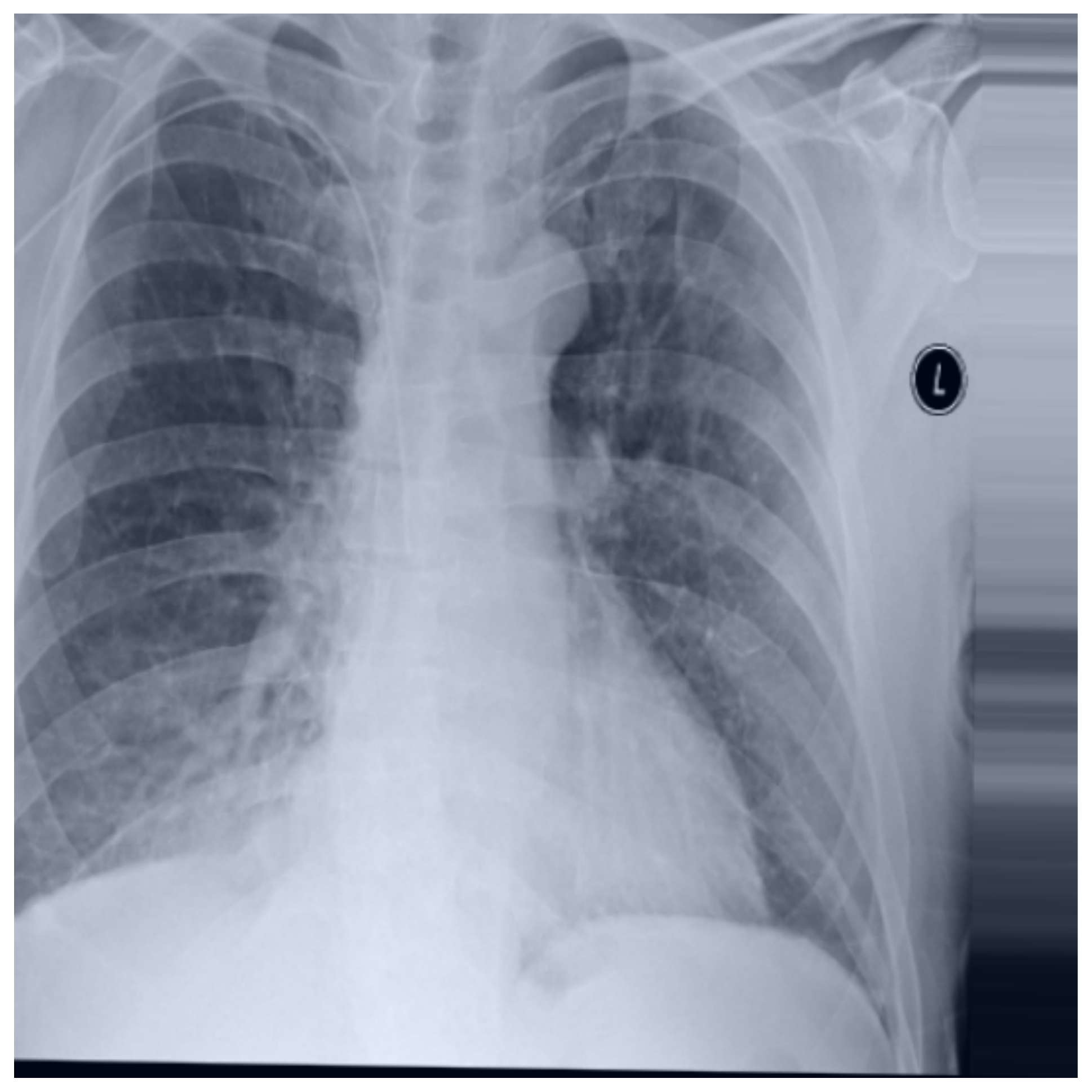}&
\includegraphics[width=1.1in,height=1.2in]{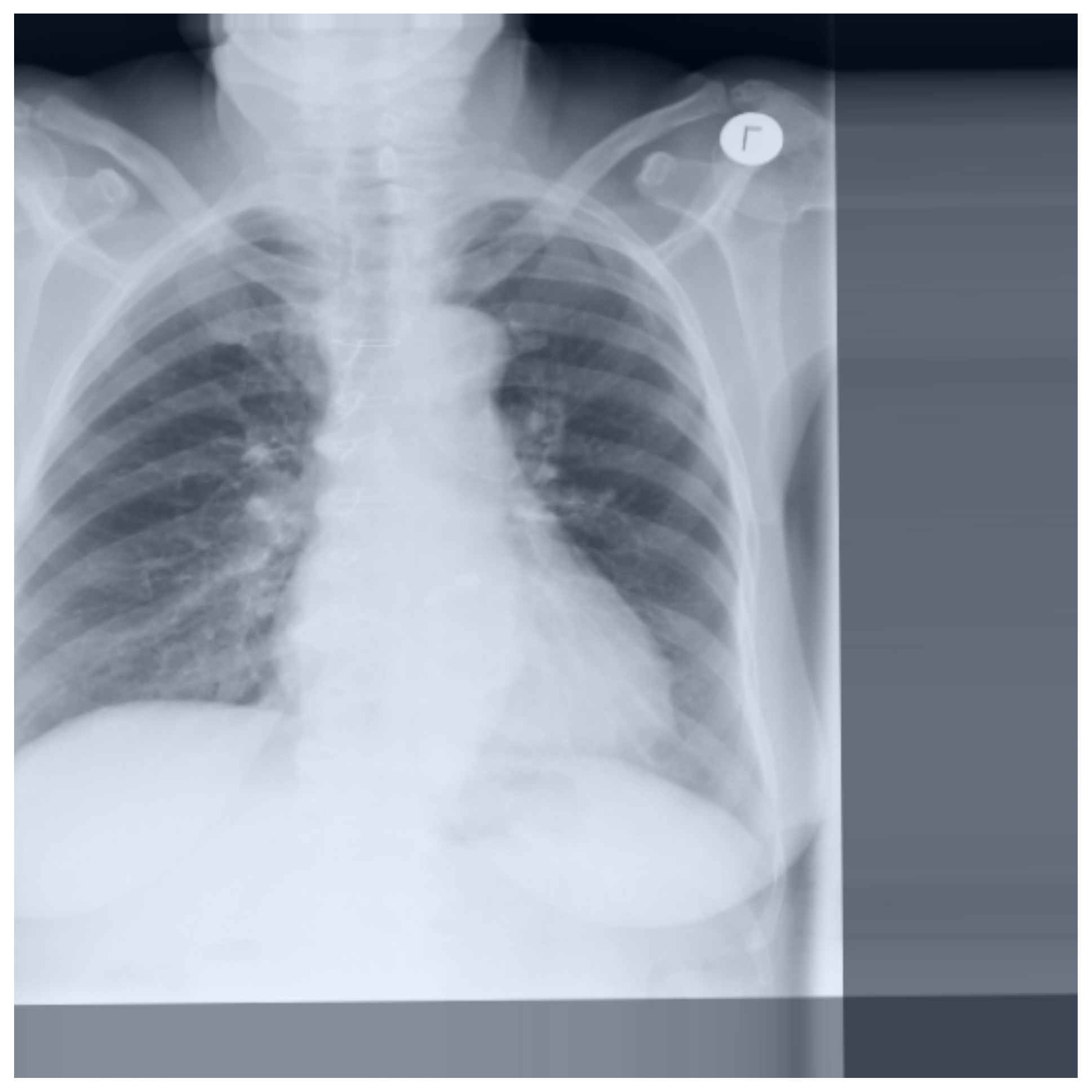}&
\includegraphics[width=1.1in,height=1.2in]{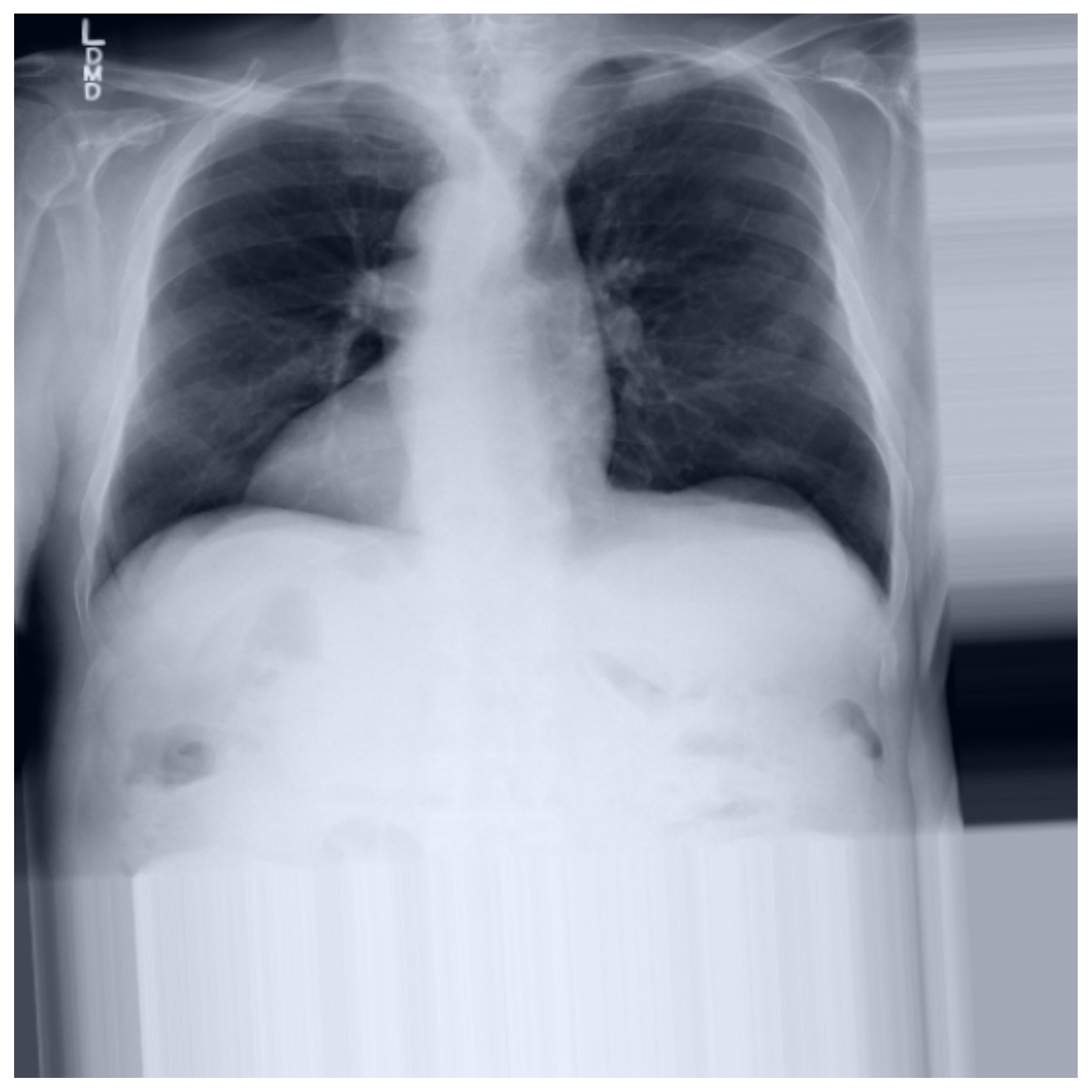}\\
\includegraphics[width=1.1in,height=1.2in]{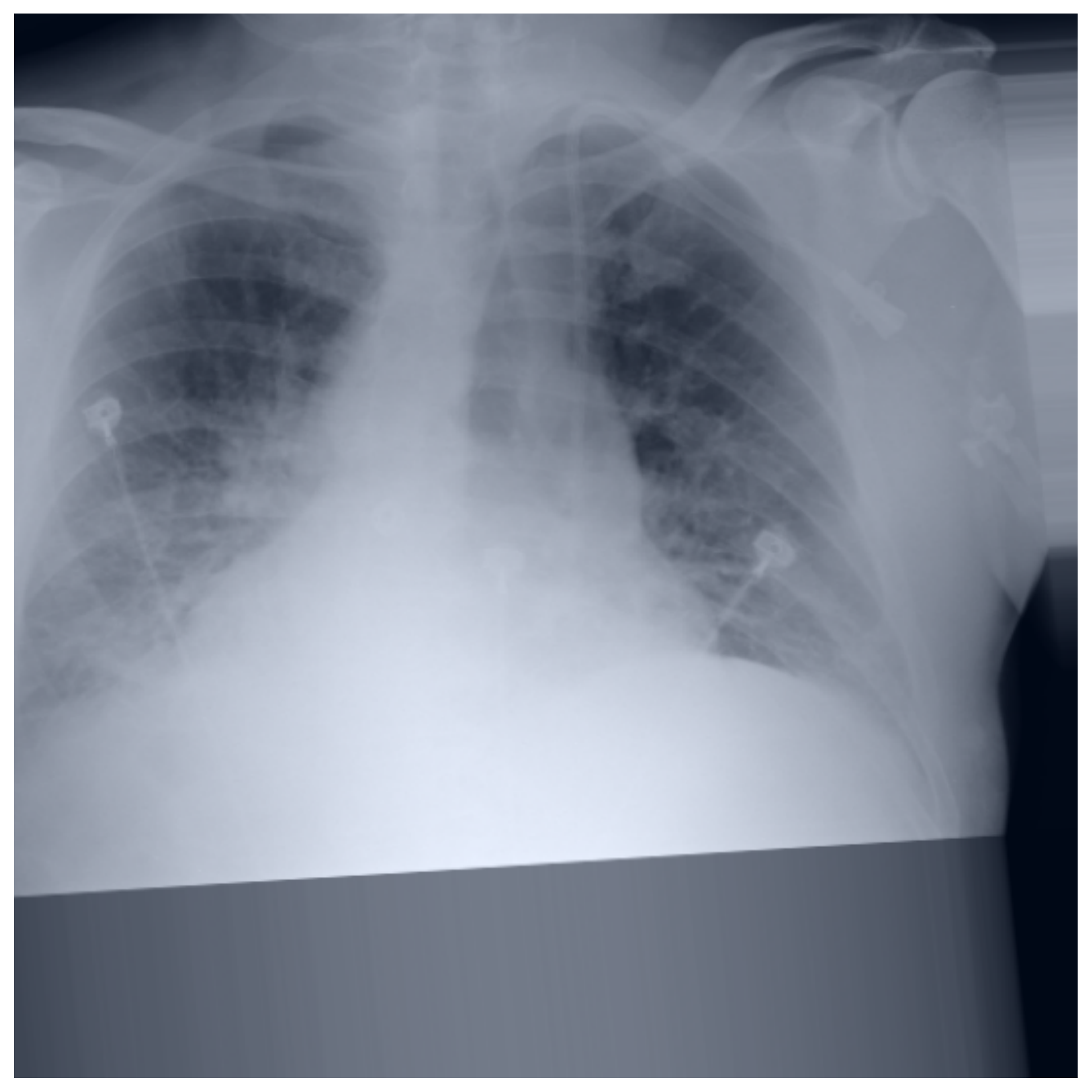}&
\includegraphics[width=1.1in,height=1.2in]{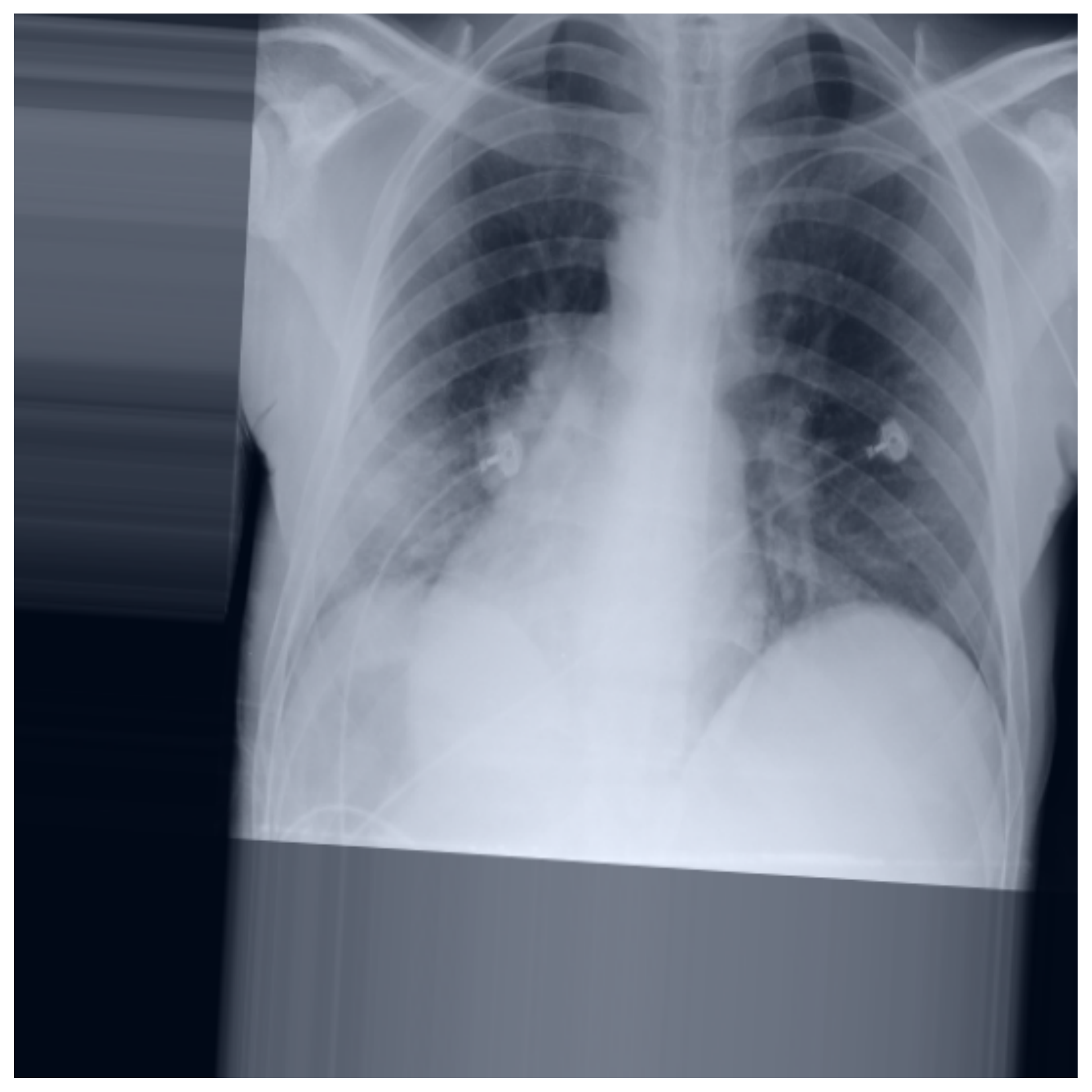}&
\includegraphics[width=1.1in,height=1.2in]{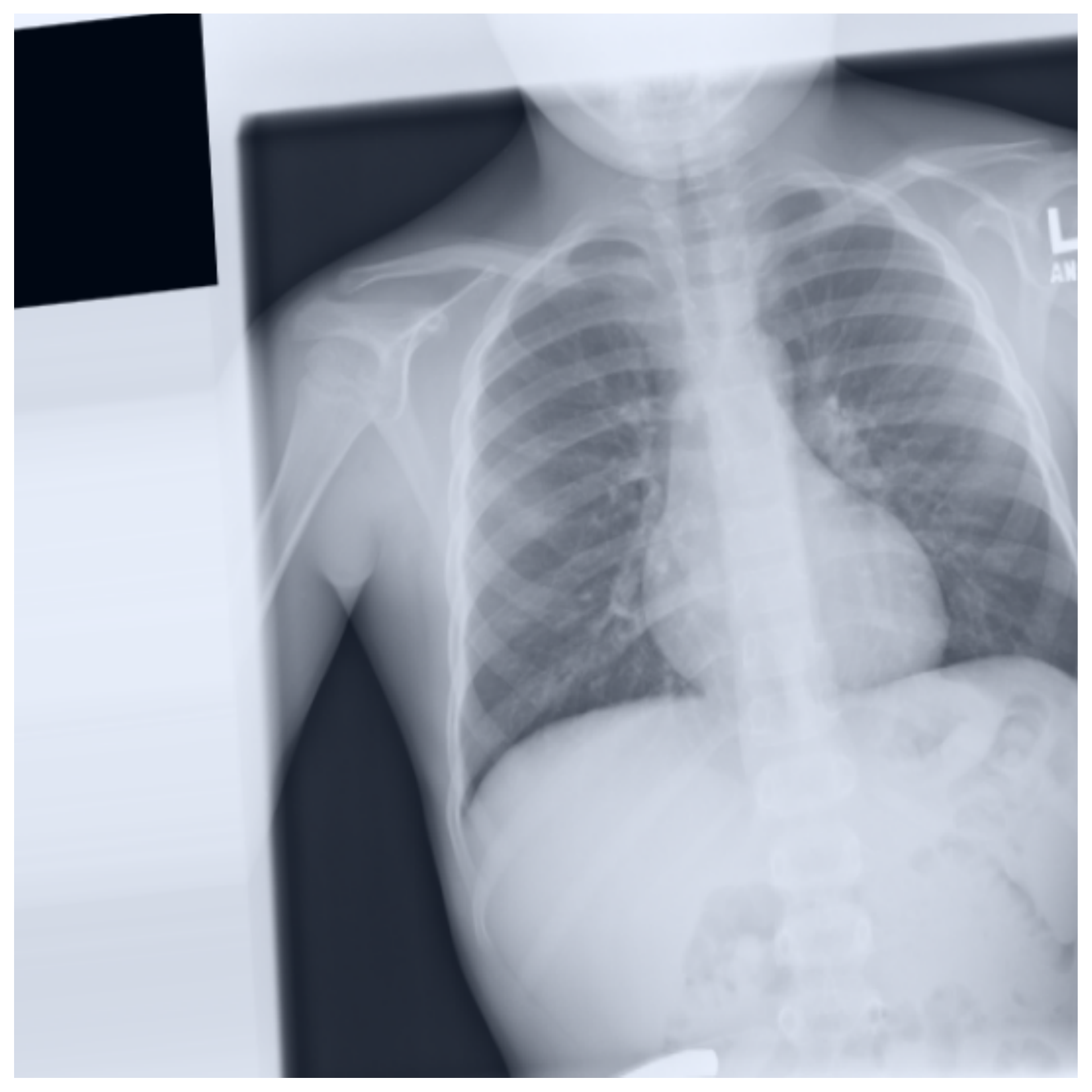}
\end{tabular}
\caption{Sample X-ray images from the NIH chest dataset.}
\label{Fig:Examples_NIH_images}
\end{figure}

In order to further assess the effectiveness of our age prediction approach, we evaluate the proposed model on the National Institutes of Health (NIH) chest X-ray dataset~\cite{wang:17}, which is comprised of 112,120 X-ray images with different ages from 30,805 unique patients. We evaluate the proposed model on 7,240 patients in the age range of 0 to 20 years old. The dataset is randomly partitioned into training (70\%), validation (10\%) and testing (20\%). Sample X-ray images from the NIH datsets are shown in Figure~\ref{Fig:Examples_NIH_images}, and the evaluation results using the MAE, RMSE and RMSPE metrics are reported in Table~\ref{Table:MAE_RMSE_and_RMSPE_Chest_dataset}.

\begin{table}[!htb]
\centering\small
\caption{Evaluation results for RidgeNet on the NIH chest X-ray test set using the MAE, RMSE and RMSPE metrics.}
\medskip
\label{Table:MAE_RMSE_and_RMSPE_Chest_dataset}
\begin{tabular}{lccc}
\toprule
Gender  & MAE     & RMSE       & RMSPE  \\ \hline
Both    & 52.21   & 61.03      & 2.56   \\
Males   & 46.61   & 55.96      & 2.75  \\
Females & 44.20   & 53.79      & 2.13  \\

\bottomrule
\end{tabular}
\end{table}

\begin{figure}[!htb]
\setlength{\tabcolsep}{.2em}
\centering
\begin{tabular}{cccc}
\includegraphics[width=1.1in,height=1.2in]{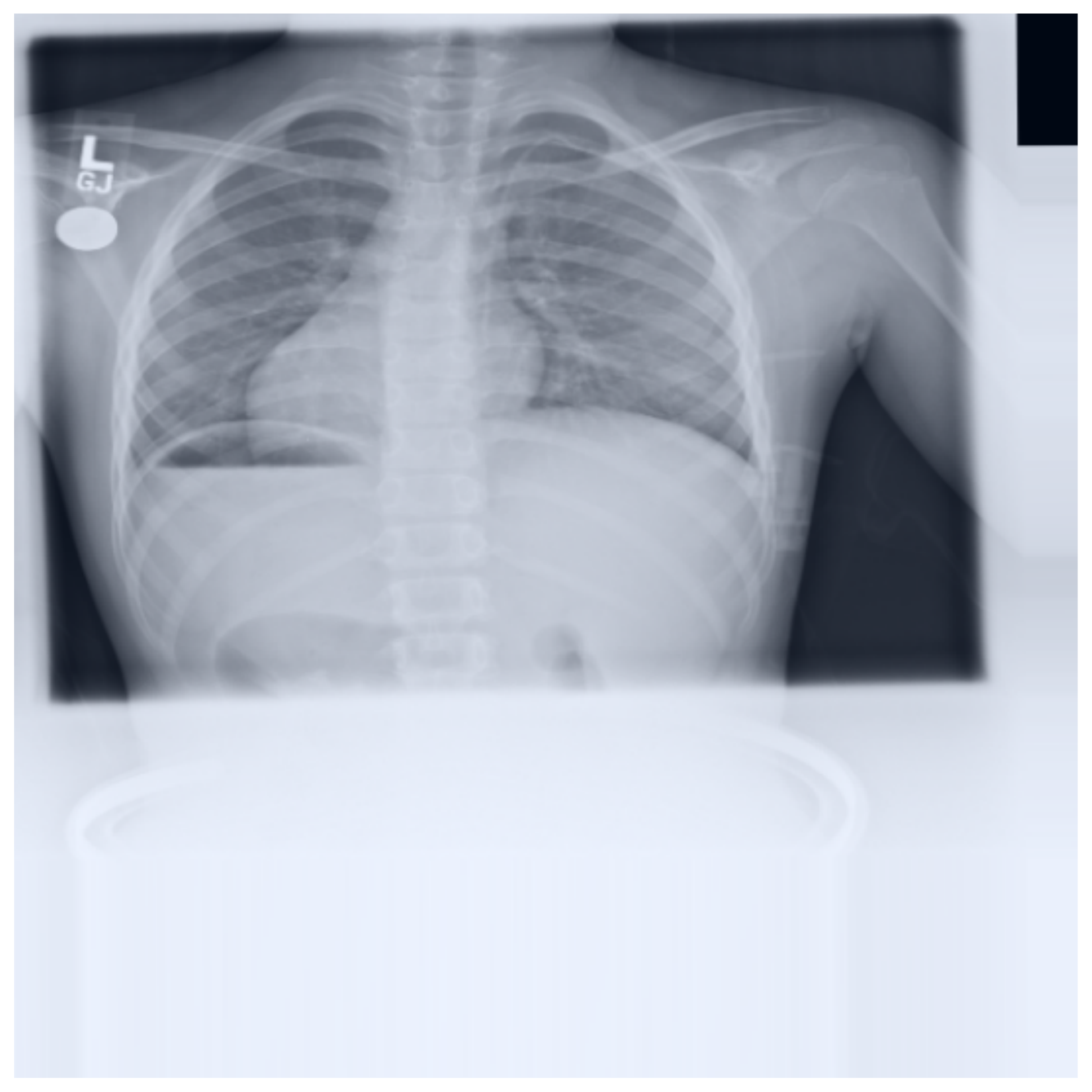}&
\includegraphics[width=1.1in,height=1.2in]{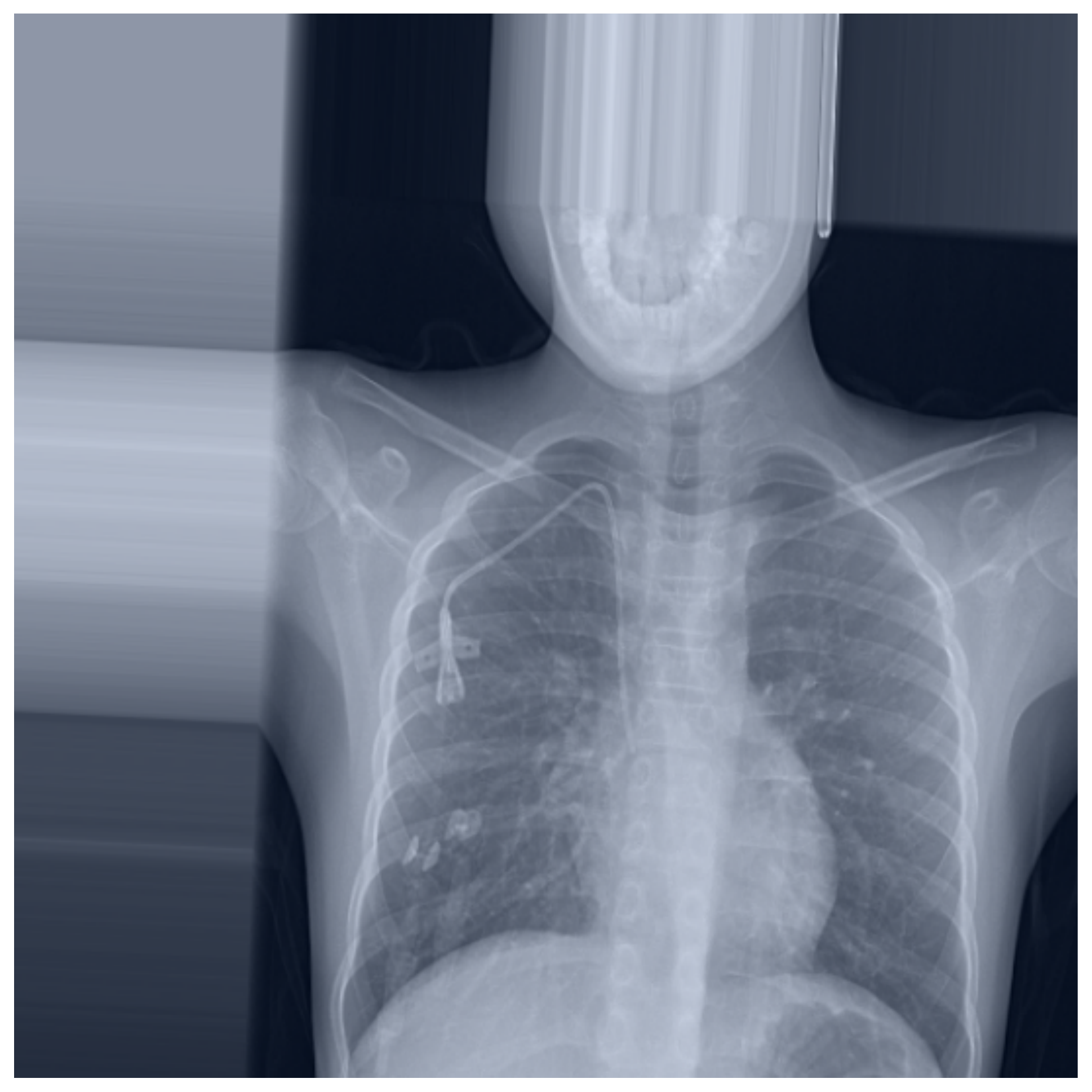}&
\includegraphics[width=1.1in,height=1.2in]{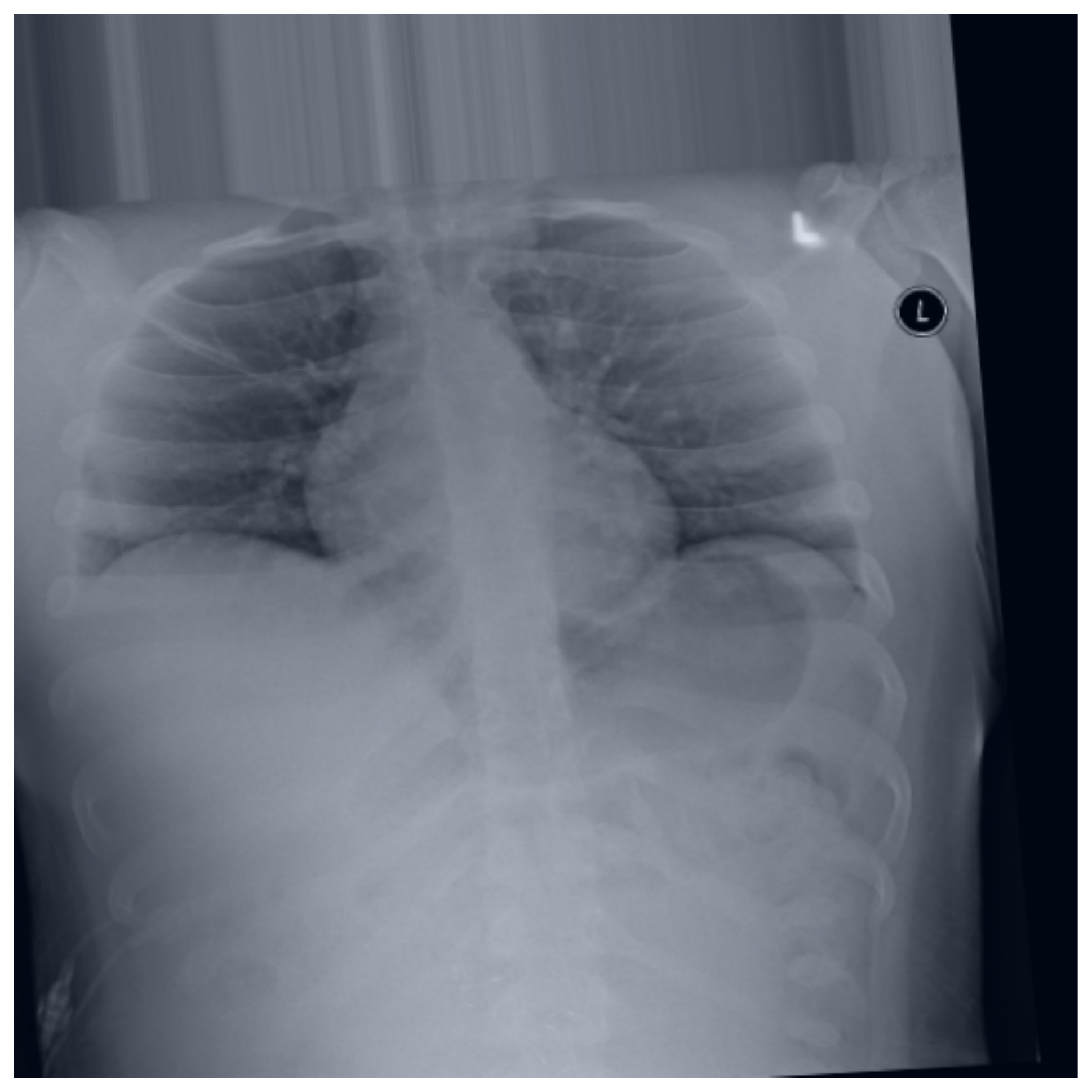}&\\
\scriptsize{Actual age: 108} & \scriptsize{Actual age: 180} & \scriptsize{Actual age: 216} \\
\scriptsize{Predicted age: 127.47} & \scriptsize{Predicted age: 138.95}  & \scriptsize{Predicted age: 143.24} \\\\
\includegraphics[width=1.1in,height=1.2in]{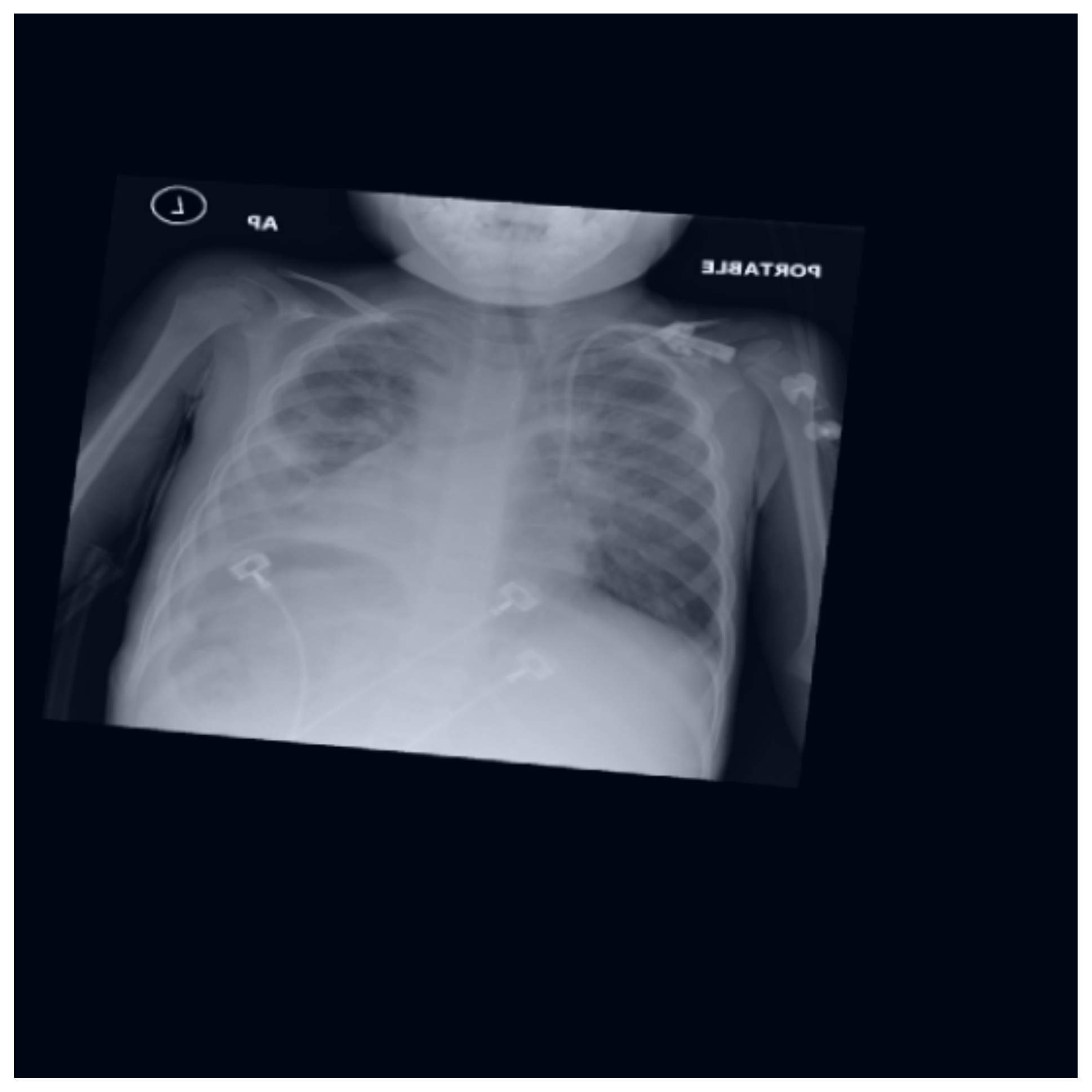}&
\includegraphics[width=1.1in,height=1.2in]{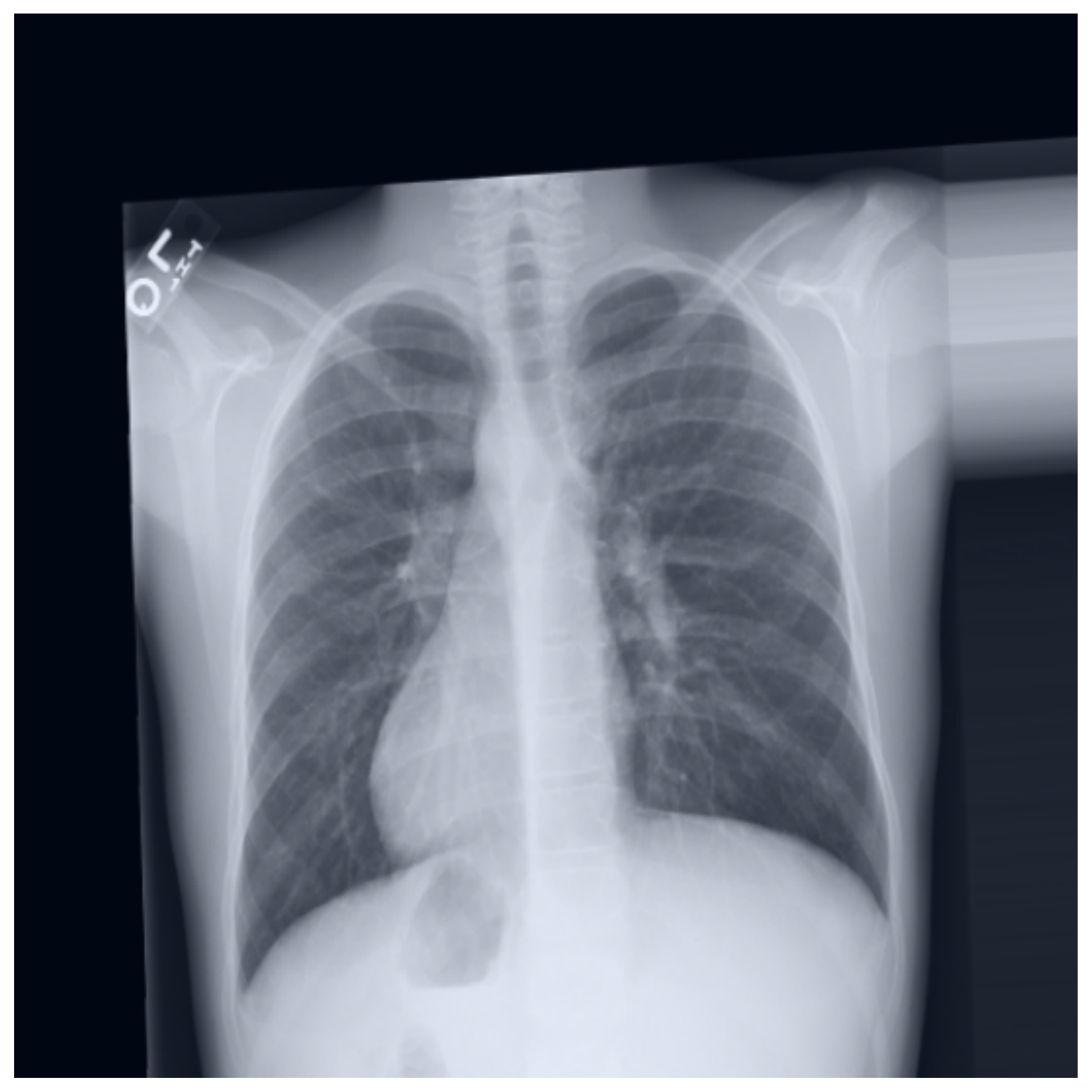}&
\includegraphics[width=1.1in,height=1.2in]{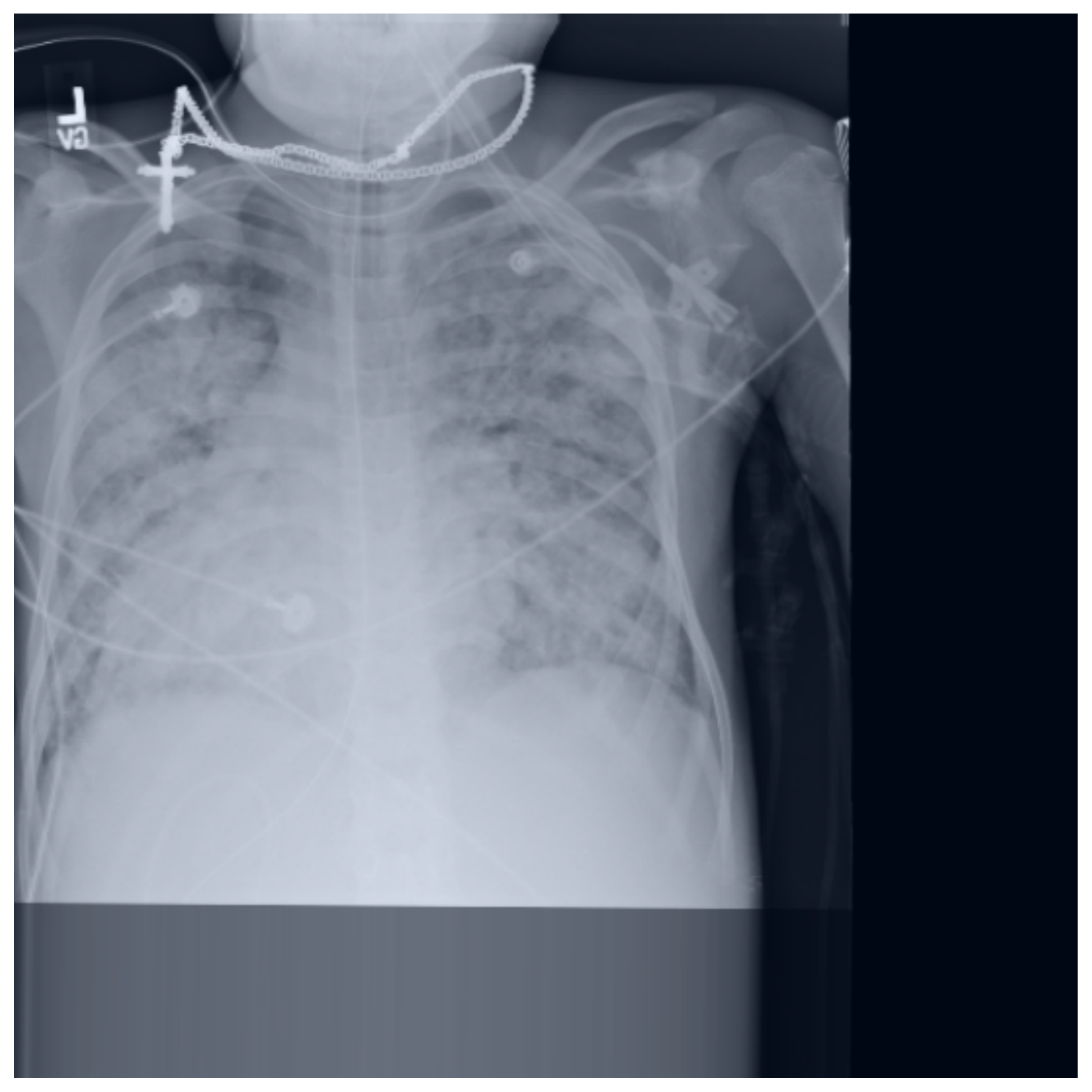}&\\
\scriptsize{Actual age: 60} & \scriptsize{Actual age: 144} & \scriptsize{Actual age: 156} \\
\scriptsize{Predicted age: 140.09} & \scriptsize{Predicted age: 160.72}  & \scriptsize{Predicted age: 129.13} \\\\
\includegraphics[width=1.1in,height=1.2in]{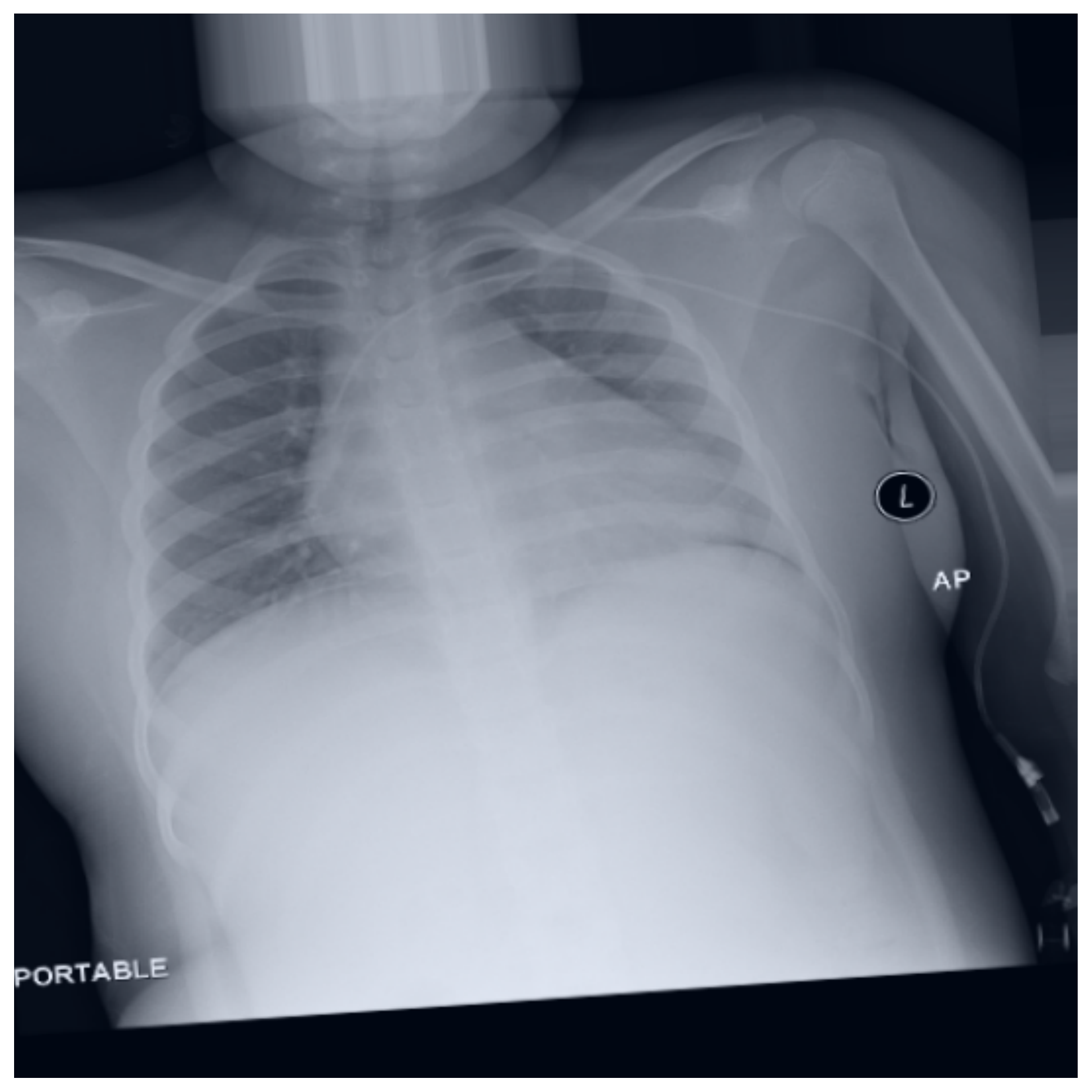}&
\includegraphics[width=1.1in,height=1.2in]{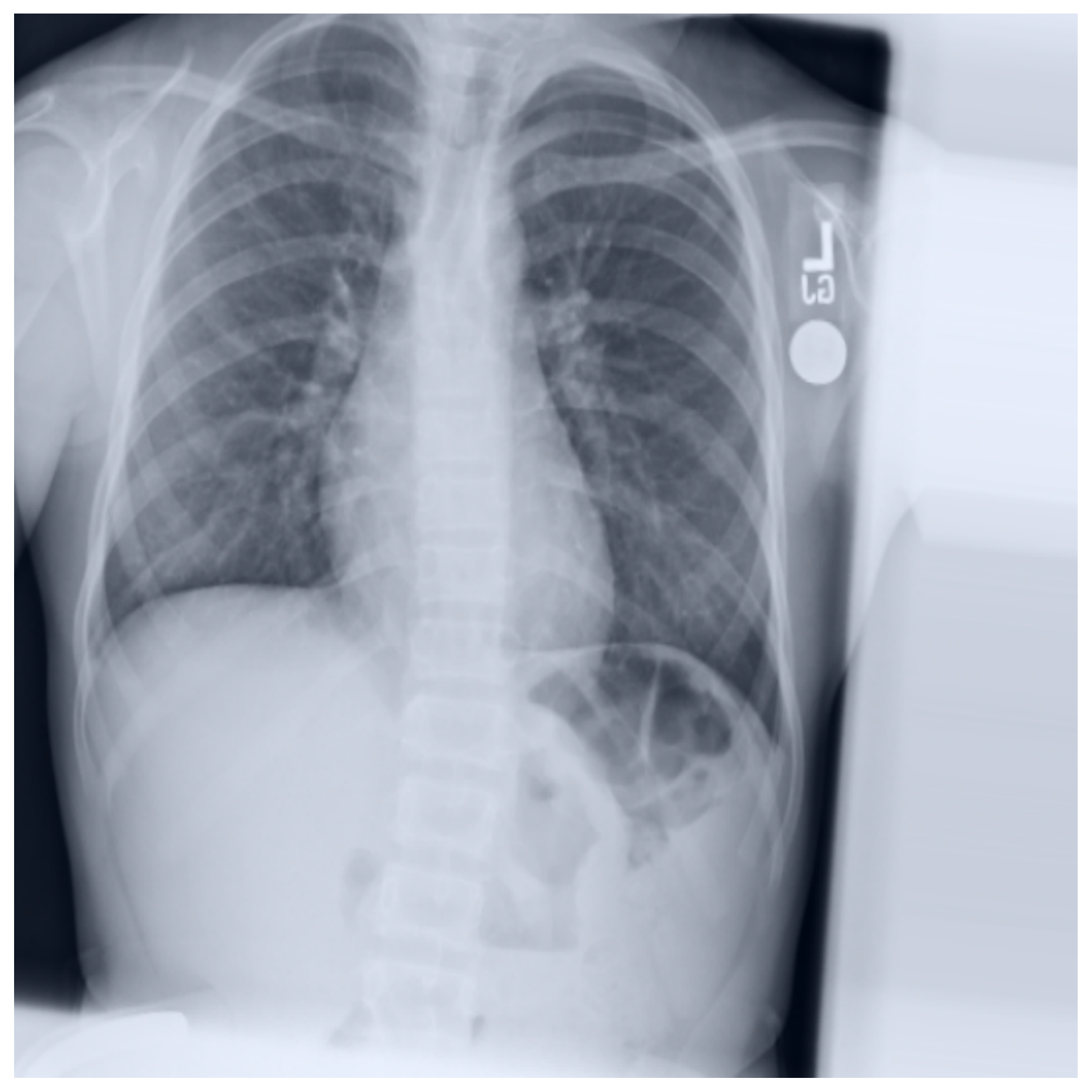}&
\includegraphics[width=1.1in,height=1.2in]{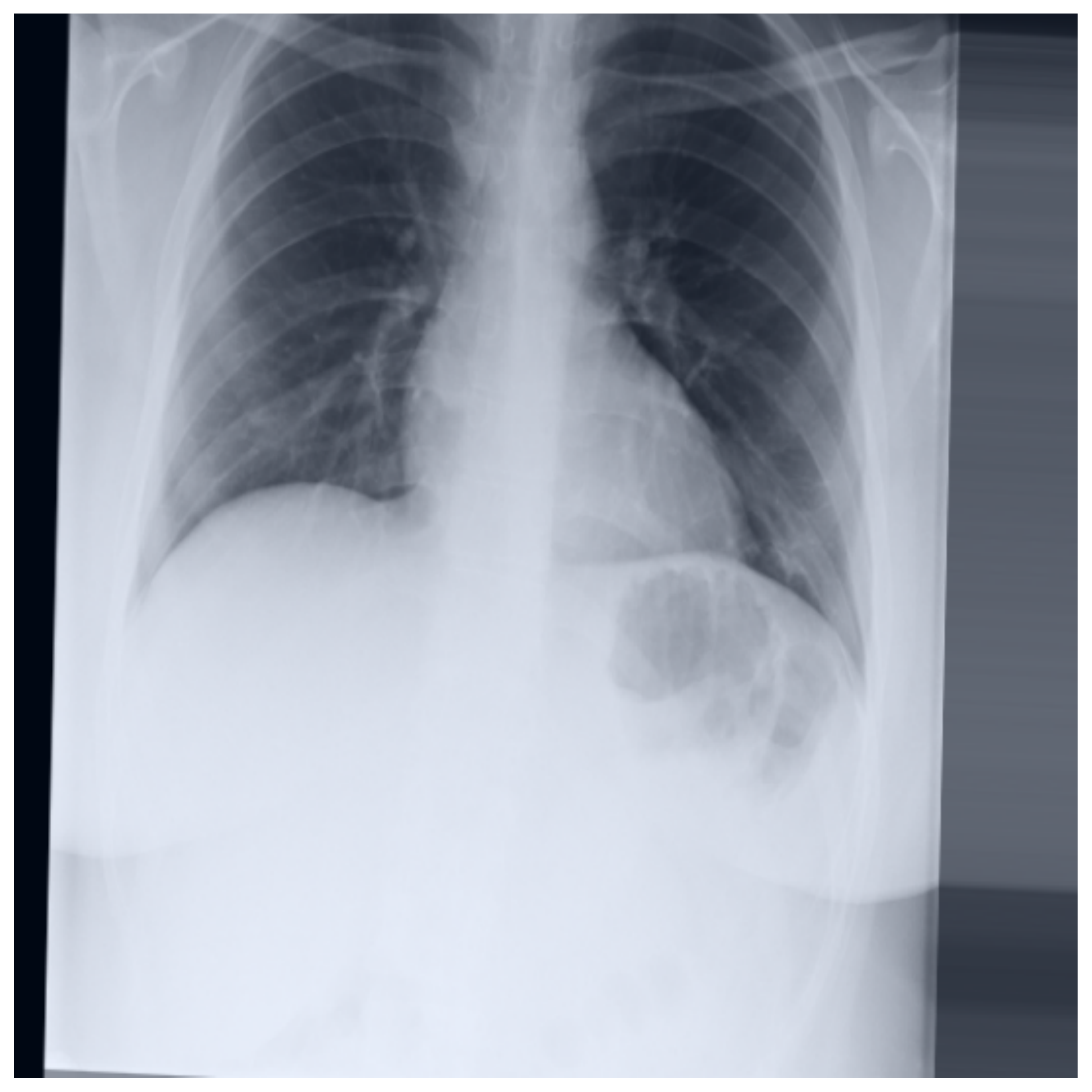}&\\
\scriptsize{Actual age: 84} & \scriptsize{Actual age: 144} & \scriptsize{Actual age: 228} \\
\scriptsize{Predicted age: 123.77} & \scriptsize{Predicted age: 144.61}  & \scriptsize{Predicted age: 179.83} \\
\end{tabular}
\caption{Actual and predicted age of sample image from the test NIH dataset in months. The X-ray images of both genders are shown in the first row, while the second and third rows display the X-ray images for males and females, respectively}
\label{Fig:Acutal_Predicted_Image_NIH}
\end{figure}

Figure~\ref{Fig:Acutal_Predicted_Image_NIH} shows the actual and predicted ages of some images from the NIH chest test set. As can be seen, the proposed model yields relatively good results, which we plan to further improve as future work.

\section{Conclusion}
In this paper, we presented a two-stage approach for bone age assessment using instance segmentation and ridge regression. The proposed framework uses instance segmentation to extract a region of interest from radiographs and background removal to avoid all extrinsic objects, followed by a regression network architecture with a ridge regression output layer that returns a single, continuous value. We also used dropout regularization to improve the generalization ability, and we shuffled the training data before each epoch to help detect overfitting issues and increase the model performance. We showed through extensive experiments on the RSNA dataset that the proposed RidgeNet model significantly outperforms existing deep learning based methods for bone age assessment, achieving the lowest mean absolute error for male and female patients, as well as for both genders. For future work, we plan to apply our model to other regression problems in medical imaging. In addition to predicting height based on the bone age, we plan to explore other network architectural designs in a bid to further improve the accuracy of the prediction results.

\bibliographystyle{ieeetr}
\bibliography{references} 

\end{document}